\documentclass[lettersize,journal]{IEEEtran}
\usepackage{amsmath,amsfonts}
\usepackage{algorithmic}
\usepackage{algorithm}
\usepackage{array}
\usepackage[caption=false,font=footnotesize]{subfig}
\usepackage{textcomp}
\usepackage{stfloats}
\usepackage{url}
\usepackage{verbatim}
\usepackage{graphicx}
\usepackage[citecolor=blue, colorlinks]{hyperref}
\usepackage{cite}
\usepackage{booktabs} 
\usepackage[dvipsnames]{xcolor}
\usepackage{tikz}

\begin{document}

\title{Enhancing Underwater Imaging with 4-D Light Fields: Dataset and Method}

\author{Yuji Lin, Junhui Hou, \textit{Senior Member}, \textit{IEEE}, Xianqiang Lyu,  Qian Zhao, and Deyu Meng, \textit{Member}, \textit{IEEE}
        % <-this % stops a space
\thanks{This project was supported in part by the NSFC Excellent Young Scientists Fund 62422118, and in part by the Hong Kong Research Grants Council under Grant 11218121, and in part by the Hong Kong Innovation and Technology Fund under Grant MHP/117/21. This work was done when Yuji Lin was at CityUHK as a visiting student. (\textit{Corresponding author: Junhui Hou and Qian Zhao})}
\thanks{Yuji Lin is with the School of Mathematics and Statistics, Xi’an Jiaotong University, Xi’an, Shaanxi 710049, China, and also with the Department of Computer Science, City University of Hong Kong, Hong Kong SAR. (email: linlos1234@gmail.com).}
\thanks{Junhui Hou and Xianqiang Lyu are with the Department of Computer Science, City University of Hong Kong, Hong Kong SAR (email: jh.hou@cityu.edu.hk; xianqialv2-c@my.cityu.edu.hk).  

Qian Zhao is with the School of Mathematics and Statistics and Ministry of Education Key Lab of Intelligent Networks and Network Security, Xi’an Jiaotong University, Shaanxi 710049, China (email: timmy.zhaoqian@mail.xjtu.edu.cn).

Deyu Meng is with the School of Mathematics and Statistics and the Ministry of Education Key Laboratory of Intelligent Networks and Network Security, Xi’an Jiaotong University, Xi’an, Shaanxi 710049, China, and also with Pazhou Laboratory (Huangpu), Guangzhou, Guangdong 510555, China (email: dymeng@mail.xjtu.edu.cn).}% <-this % stops a space
}

\markboth{IEEE Journal of Selected Topics in Signal Processing}%
{Shell \MakeLowercase{\textit{et al.}}: A Sample Article Using IEEEtran.cls for IEEE Journals}


\maketitle

\begin{abstract}
In this paper, we delve into the realm of 4-D light fields (LFs) to enhance underwater imaging plagued by light absorption, scattering, and other challenges. 
Contrasting with conventional 2-D RGB imaging, 4-D LF imaging excels in capturing scenes from multiple perspectives, thereby indirectly embedding geometric information. This intrinsic property is anticipated to effectively address the challenges associated with underwater imaging. 
By leveraging both explicit and implicit depth cues present in 4-D LF images, we propose a progressive, mutually reinforcing framework for underwater 4-D LF image enhancement and depth estimation. 
The entire framework decomposes this complex task, iteratively optimizing the enhanced image and depth information to progressively achieve optimal enhancement results. 
More importantly, we construct the first 4-D LF-based underwater image dataset for quantitative evaluation and supervised training of learning-based methods, comprising 75 underwater scenes with multiple views and 3675 high-resolution 2K pairs.
To craft vibrant and varied underwater scenes, we build underwater environments with various objects and adopt several types of degradation. Through extensive experimentation, we showcase the potential and superiority of 4-D LF-based underwater imaging vis-a-vis traditional 2-D RGB-based approaches. Moreover, our method effectively corrects color bias and achieves state-of-the-art performance. The dataset and code will be publicly available at \href{https://github.com/linlos1234/LFUIE}{https://github.com/linlos1234/LFUIE}.
\end{abstract}


\section{Introduction}
\IEEEPARstart{U}{nderwater} imaging is one of the key technologies for exploring the marine world. It plays a vital role in studying marine life and ecosystems, helping us to gain a deeper understanding of the complex underwater environments. However, the quality of underwater imaging is often affected by various degradation factors \cite{akkaynak2017space}. These include the selective absorption of light, which depends on the distance photons travel, and the scattering caused by light interacting with numerous particles suspended in the water. As a result, captured underwater images often suffer from color cast, low contrast, and noise. Therefore, effectively harnessing mechanisms behind underwater degradation is essential for enhancing imaging.

\begin{figure}[!t]
\centering
\includegraphics[width=1\linewidth]{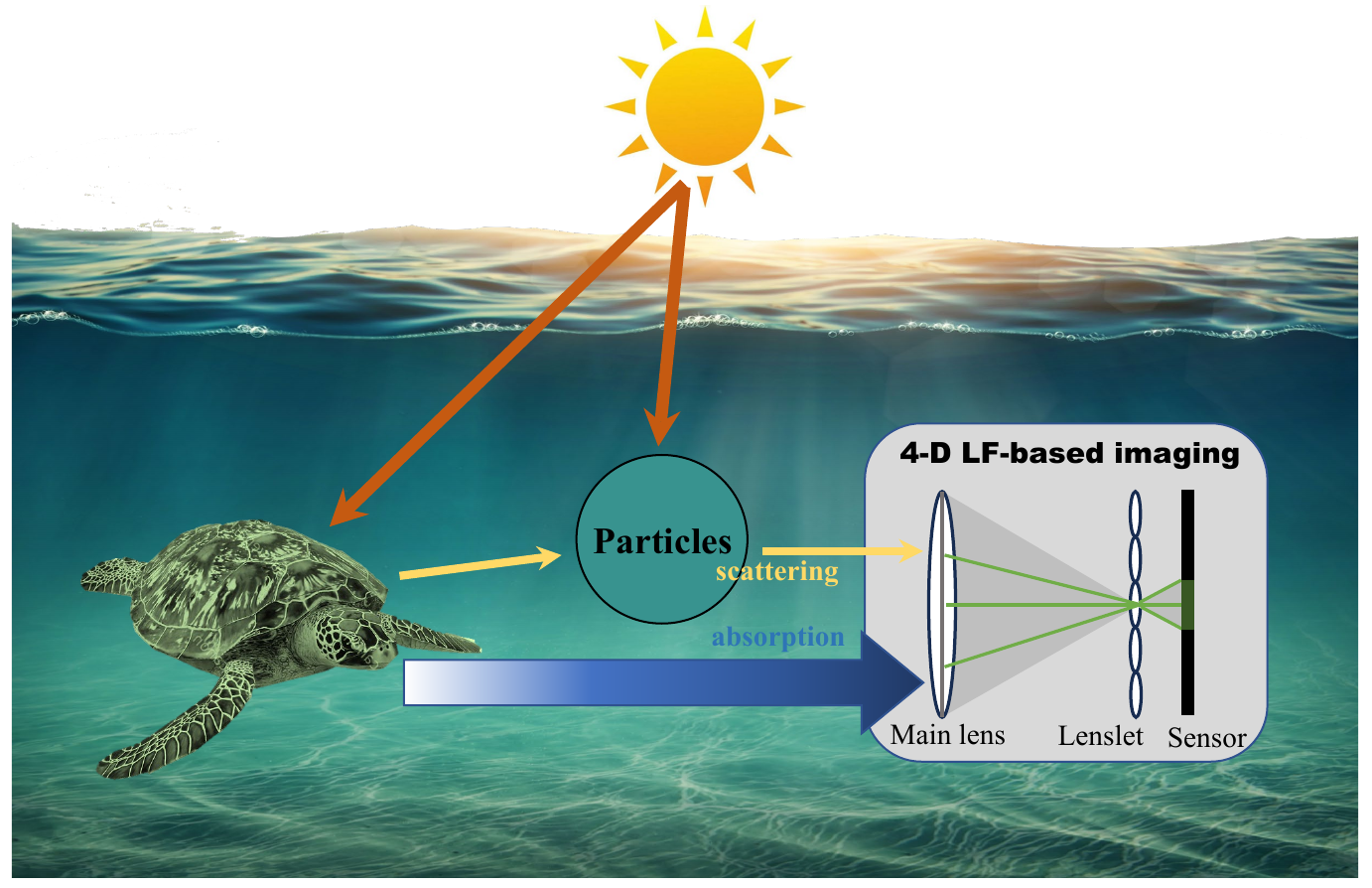}
\caption{We explore the advanced 4-D LF-based technique to enhance underwater imaging suffering from absorption and scattering. The multi-view property and implicitly encoded geometry information of scenes (i.e., depth information) by 4-D LFs are expected to mitigate these degradations.}
\label{fig:insight}
\vspace{-4mm}
\end{figure}

Over the past years, many physical model-based and learning-based enhancement methods \cite{chiang2011underwater, li2016underwater1, li2016underwater2, berman2020underwater, li2019underwater, liu2022adaptive} have been proposed for improving the quality of underwater images obtained by traditional 2-D RGB imaging. These physical model-based approaches \cite{chiang2011underwater, li2016underwater1, li2016underwater2, berman2020underwater} mainly concentrate on the depth-related parameters to the degradation process of the underwater imaging, while the deep learning algorithms \cite{li2019underwater, liu2022adaptive} apply the popular convolutional or transformer network architecture to overfit 2-D RGB-based datasets. However, 
these traditional 2-D RGB-based underwater imaging approaches still face many challenges, including complex and ill-posed optimization for model-based methods and the lack of realistic pairwise data for learning-based methods. Moreover, monocular depth estimations from 2-D RGB imaging are often inaccurate, causing limited performance.

4-D light field (LF) imaging is an advanced multi-dimensional technique \cite{levoy1996light, ng2005light} that can capture the same scene from multiple perspectives in a single shot. This technology can simultaneously record both the reflected light intensity and rich geometric information of the scene. Under the assumption of Lambertian, projections of the same scene point have the same intensity across different views. This geometric relation leads to a particular LF parallax structure, which makes depth estimation easier compared to traditional 2-D RGB imaging. In addition, LF images contain implicit geometric information, enabling the aggregation of valuable content from all views. By aligning features, images from different perspectives can merge light rays and reduce noise while preserving geometric structures, resulting in improved consistency. Therefore, it is promising to explore the acquisition of high-quality underwater images through 4-D LF imaging. 

\begin{figure*}[!t]
\centering
\includegraphics[width=1\linewidth]{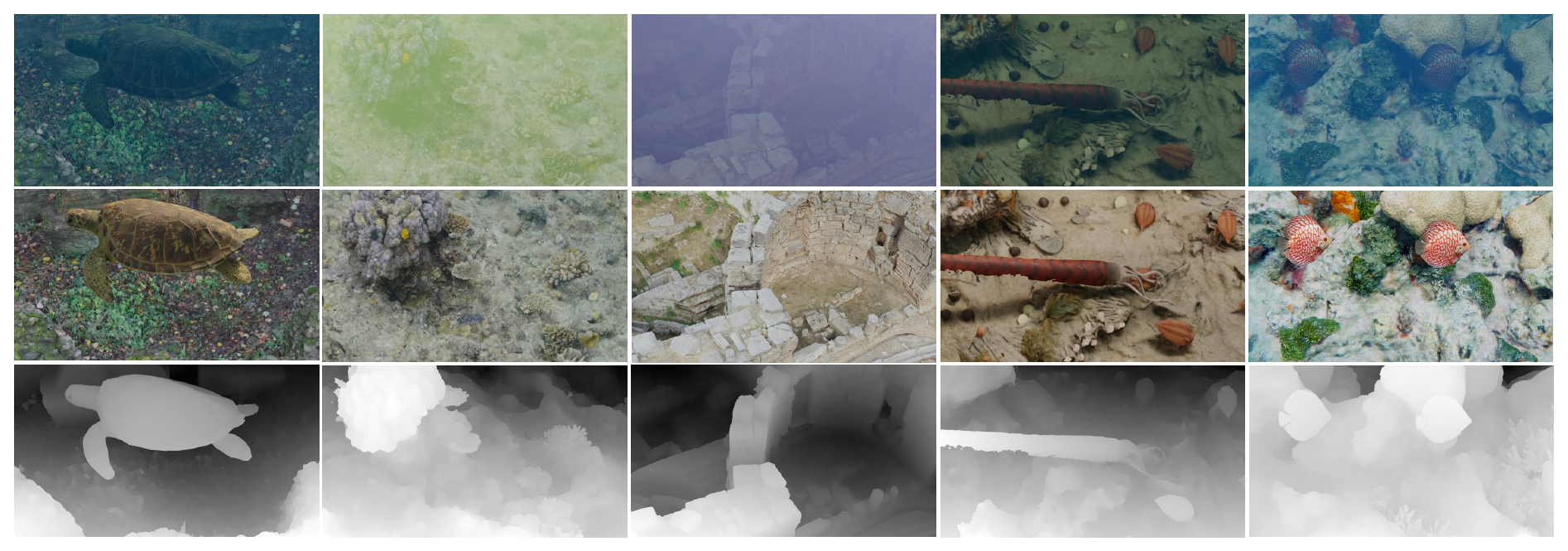}
\caption{Samples from our LFUB dataset with different color deviation degradations. Top row: central views of underwater 4-D LF images.  Middle row: the corresponding ground truth images; Bottom row: the corresponding ground truth depth maps.}
\label{fig:sample-images}
\vspace{-4mm}
\end{figure*}

4-D LF images captured in underwater environments are essential for exploring this potential direction. Unfortunately, there is currently no publicly available dataset for such imagery. While LF cameras could be employed for data acquisition, this approach is laden with challenges, such as high expenses and the absence of authentic ground truth data. This scarcity complicates the quantitative assessment of LF technology's potential in underwater imaging and also hinders supervised learning of networks. Consequently, we opt to synthesize lifelike underwater 4-D LF images through 3-D modeling and rendering software. Specifically, we utilize \textit{Blender} \cite{blender} to conduct the first 4-D LF-based underwater image dataset that includes LF images, ground truth images, and depth maps, as depicted in Fig. \ref{fig:sample-images}. Furthermore, we propose a learning-based method to enhance underwater 4-D LF images. Our key innovation lies in that depth estimation and enhancement results promote each other. The method enhances 4-D LF images by iterating multiple stages with both explicit and implicit geometric information. Specifically, we utilize the estimated depth maps from 4-D LF images to modulate the features. The geometric information extracted from epipolar plane image (EPI) features can also help adjust the spatial and angular features. Extensive experiments and ablative studies demonstrate the advantages of our approach. 

In summary, the key contributions of this paper are:
\begin{itemize}
    \item we innovatively introduce 4-D LF-based underwater imaging, in which the inherent geometric and multi-view information of 4-D LF images could potentially facilitate the subsequent enhancement process for producing high-quality underwater images; 
    \item we build the \textit{first} 4-D LF-based underwater image dataset LFUB, which contains 75 multi-view scenes of realistic and high-resolution underwater 4-D LF images with corresponding no-water LF images and depth maps. Our LFUB dataset is expected to advance research in the field of underwater vision, and the pipeline of conducting is feasibly used to build other helpful LF-based datasets; 
    \item we propose a learning-based method, which leverages the unique characteristics of 4-D LFs progressively to enhance the quality of 4-D LF images; and 
    \item we conduct a comprehensive evaluation of various existing algorithms related to image, video, and LF processing. The insights derived from these evaluations hold promise for the broader community.
\end{itemize}

The remainder of this paper is organized as follows: Section \ref{sec-RW} briefly reviews the line of related works in the field. Section \ref{sec-dataset} details our LFUB dataset. Section \ref{sec-method} provides a detailed introduction to our proposed framework, followed by extensive experiments and analyses in Section \ref{sec-exp}. Finally, Section \ref{sce-c} concludes this paper and discusses some directions for improving the proposed dataset and algorithm.

\section{Related work}
\label{sec-RW}
\subsection{Underwater 2-D RGB Image/Video Enhancement}
The field of underwater 2-D RGB image enhancement has developed through a variety of methodologies aimed at correcting distortions caused by the optical properties of water. These approaches are categorized into physical model-free, physical model-based, and data-driven methods.

\emph{1) Physical Model-free Methods.} These approaches enhance the perceived visual quality by modifying pixel values directly without the use of physical models. Techniques include: histogram equalization \cite{hitam2013mixture}, image fusion \cite{ancuti2012enhancing, ancuti2017color} and Retinex-based methods \cite{fu2014retinex, zhang2017underwater, zhuang2021bayesian}. For instance, Hitam \emph{et al.} \cite{hitam2013mixture} introduced a hybrid method that enhanced underwater images by integrating the strengths of contrast adjustment and histogram equalization across both the RGB and HSV color spaces. Ancuti \emph{et al.} \cite{ancuti2012enhancing} improved image quality by combining different enhanced versions of inputs through a weighted and multi-scale fusion strategy. Zhuang \emph{et al.} \cite{zhuang2021bayesian} developed a Bayesian Retinex algorithm, which simplified underwater image enhancement into two straightforward sub-problems with an efficient optimization algorithm.

\emph{2) Physical Model-based Methods.} Widely used physical model-based methods treat underwater image enhancement as an inverse problem, which utilizes prior knowledge to estimate the parameters of underwater imaging models. Researchers typically improve underwater images through the following sequential steps: defining the initial physical imaging model, estimating crucial parameters, and solving the inversion problem. The foundation of these methods lies in designing appropriate priors based on a physical understanding of the underwater world to accurately characterize the environment. These priors involve dark channel prior \cite{he2010single}, red channel prior \cite{galdran2015automatic}, minimum information prior \cite{li2016underwater1, li2016underwater2}, haze-line prior \cite{berman2020underwater}, etc.

Carlevaris-Bianco \emph{et al.} \cite{carlevaris2010initial} noted that red light often attenuates faster than green and blue light in underwater environments. Based on this observation, they designed a prior that leverages this differential attenuation to estimate transmission maps of underwater images. Drews \emph{et al.} \cite{drews2013transmission} proposed the Underwater Dark Channel Prior (UDCP) to address the selective attenuation of light. Li \emph{et al.} \cite{li2016underwater1} enhanced underwater images by employing a dehazing technique that minimizes information loss, incorporating a histogram distribution prior to improve image details. In addition to the minimum information prior, Li \emph{et al.} \cite{li2016underwater2} considered the specific optical properties of underwater environments. Berman \emph{et al.} \cite{berman2020underwater} adapted the haze-line prior for underwater images, which is based on the concept that different wavelengths are absorbed at different rates, causing colors to shift primarily towards blue and green. 
Zhuang \emph{et. al} \cite{zhuang2022underwater} proposed a retinex variational model using hyper-Laplacian reflectance priors, which applied sparsity-promoting and complete-comprehensive reflectance.
Moreover, Akkaynak and Treibitz \cite{akkaynak2019sea} revised the underwater image formation model \cite{akkaynak2018revised} and presented Sea-Thru, a physically accurate color correction method. 

\emph{3) Data-Driven Methods.} In recent years, deep learning has significantly progressed in addressing low-level vision tasks \cite{guo2020zero, wang2020model, fu2022kxnet}. Underwater image enhancement has particularly benefited from the powerful fitting capabilities of deep learning methods, coupled with massive synthetic training data, achieving impressive performance.

Initially, due to the scarcity of pairwise data, Li \emph{et al.} \cite{li2017watergan} employed WaterGAN and an imaging formation model to synthesize pairwise data with clean images and depth maps in an unsupervised manner. Subsequently, they adopted a two-stage training strategy to restore underwater images, especially for color correction. Li \emph{et al.} \cite{li2018emerging} proposed a weakly supervised method, Water CycleGAN based on Cycle Consistent Adversarial Networks \cite{zhu2017unpaired} and a multi-term loss function, relaxing the need for pairwise training data of underwater images. More recently, Li \emph{et al.} \cite{li2019underwater} introduced the first pairwise underwater image dataset, which greatly advanced the performance of supervised learning and proposed WaterNet as a baseline network.
Liu \emph{et al.} \cite{liu2022adaptive} designed an adaptive learning module to learn representative features and transfer shallow information by residual learning.
Uncertain learning \cite{fu2022uncertainty} was utilized to help improve enhancement performance.
Guo \emph{et al.} \cite{guo2023underwater} pre-trained a ranking-based model using a histogram prior and cross-scale correspondence to determine which input is better, then trained a simple enhancement network with pre-trained rank loss to achieve optimal outputs.
Peng \emph{et al.} \cite{peng2023u} used a UNet shape transformer network with channel-wise and spatial-wise feature modulation and proposed a multi-color space loss function to improve image contrast. 
Wang \emph{et al.} \cite{wang2023domain, wang2024uierl} enhanced the underwater image quality by mining internal and external information and using domain adaptation and representation learning methods. 
Moreover, Li \emph{et al.} \cite{li2021underwater} used physical transmission maps as guidance to modulate features and exploited information from multiple color spaces to improve image quality.

In addition to single image methods, there are also some approaches for video-based underwater enhancement \cite{du2024end,xie2024uveb}, which leverage additional temporal information for improved results. Xie \emph{et al.} \cite{xie2024uveb} exploited the inter-frame information to optimize the enhancement guided by the information from the current frame.

\subsection{Light Field Image Restoration}
Researchers developed different network architectures to represent various geometric information of LF images for efficient LF image enhancement and restoration.
Lamba \emph{et al.} \cite{lamba2020harnessing} employed the intrinsic multi-view capabilities of LF to improve images under low-light conditions. Lamba \emph{et al.} \cite{lamba2022fast} proposed a fast and resource-efficient approach to restore extremely dark LF images. Zhang \emph{et al.} \cite{zhang2023lrt} developed a transformer-based model tailored for low-light conditions. Wang \emph{et al.} \cite{wang2023multi} introduced a novel framework using multi-stream networks to progressively enhance and denoise low-light LF images. Lyu \emph{et al.} \cite{lyu2024enhancing} proposed a more interpretable unfolding network to leverage the characteristics of low-light LF images. Chen \emph{et al.} \cite{chen2018light} designed a novel LF denoising framework based on anisotropic parallax analysis, which jointly trained two convolutional neural networks to extract inter-view and view-specific features. Guo \emph{et al.} \cite{guo2021deep} adopt an implicit regularization term to embed the LF structure prior. Lyu \emph{et al.} \cite{lyu2024probabilistic} designed to learn a feature embedding architecture by assembling various low-dimensional convolution patterns in a probability space for fully capturing spatial-angular information. Ding \emph{et al.} \cite{ding2021rain} introduced a GAN-based architecture that uses depth information to eliminate rain streaks from 3-D EPIs of rainy LF images. Yan \emph{et al.} \cite{yan2022rain} applied 4-D convolutions along with multi-scale Gaussian processes for effective rain removal from LF images. Lyu \emph{et al.} \cite{lyu2024rainyscape} proposed an unsupervised paradigm to remove rain streaks from multi-view rainy images using neural rendering. In addition to low-light enhancement, noise removal, and rain removal tasks, there is no previous work on enhancing LF-based underwater imaging.

\begin{table*}[t]
\scriptsize
  \caption{Comparison of pairwise underwater datasets. The data simulation methods include underwater image synthesis algorithm \cite{akkaynak2019sea}, GAN-based algorithm \cite{fabbri2018enhancing}, enhancement method \cite{li2019underwater}, and commercial software (Blender). Note that the enhancement-based methods contain real underwater images and synthetic no-water images}
  \label{tab-dataset}
  \centering
  \small
  \setlength{\tabcolsep}{1.2mm}{
  \begin{tabular}{c|ccccc}
    \toprule[1.2pt]
    ~~~~Dataset~~~~
    &~~~Data Type~~~&~~~Spatial Resolution~~~&~~~Amount~~
    &~~~Modality~~~&~~~Acquired Method~~~ \\
    \midrule
     UGAN \cite{fabbri2018enhancing}    
     &Syn &256 $\times$ 256 &8,000 &RGB Image  &GAN-based       \\
     UIEB \cite{li2019underwater}
     &Syn/Real &299 $\times$ 168 &890 & RGB Image  &Enhanced \\
     EUVP \cite{islam2020fast}
     &Syn &256 $\times$ 256  &12,000 &RGB Image &GAN-based   \\
     % LNRUD \cite{ye2022underwater}
     % &Syn &256$\times$ 256  &5k &single &Neural Rendering   \\ 
     Syrea \cite{wen2023syreanet}
     &Syn &256 $\times$ 256  &20,000 &RGB Image &Imaging Model   \\ 
     LSUI \cite{peng2023u}
     &Syn/Real  &256 $\times$ 256  &4,000 &RGB Image &Enhanced   \\  
     SUVE \cite{du2024end}
     &Syn &170$\times$ 256$\times$ 256  &660 &RGB Video &GAN-based   \\  
     UVEB \cite{xie2024uveb}
     &Syn/Real &180 $\times$ 960$\times$ 528  &1308 &RGB Video &Enhanced  \\
     LFUB
     &Syn &7 $\times$ 7 $\times$ 1920 $\times$ 1080  &75 &4-D Light Field &Blender Rendering
     \\
     \bottomrule[1.2pt]
  \end{tabular}}
\vspace{-4mm}
\end{table*}

\subsection{Underwater 2-D RGB Image/Video Dataset}
The development of underwater datasets has recently garnered significant attention. To analyze comparisons, we provide a summary of frequently used 2-D RGB image/video-based datasets in Table \ref{tab-dataset}, which includes details on data resolution, availability of ground truth, modality as well as acquisition method. 

Acquiring pairwise real-world underwater image datasets is challenging due to the complexities of underwater environments and the difficulties in obtaining accurate underwater ground truth images (GTs). As a result, researchers have conducted synthetic underwater image datasets using a revised imaging model \cite{akkaynak2019sea} and a GAN-based method \cite{fabbri2018enhancing}. Wen \emph{et al.} \cite{wen2023syreanet} synthesized pairwise underwater/no-water images by applying empirical parameters from the revised imaging model \cite{akkaynak2019sea} to indoor images. Additionally, \cite{fabbri2018enhancing, islam2020fast} leveraged the GAN model to learn the underwater image distribution and then sampled from this distribution to generate pairwise underwater/no-water images. However, these methods do not accurately represent the real underwater world, as they fail to adequately capture the effects of underwater scattering, absorption, and noise. Alternatively, the popular dataset \cite{li2019underwater} utilized various released methods to enhance raw images and invited volunteers to vote for the best results as GTs. While this strategy provides access to real underwater images, the GTs are still produced by other model-free approaches or data-driven enhancement methods trained on synthesized datasets. Thus, the distribution learned from the enhancement-based dataset tends to overfit a sub-optimal manifold influenced by the distributions of previous synthetic datasets.

Overall, recent underwater image datasets have not directly obtained both synthetic underwater images and synthetic GTs. Furthermore, there is no mature strategy for applying these types of methods to LF images. Specifically, traditional 2-D RGB-based approaches compromise the consistency of inter-view data when directly applied, leading to the lack of coherence between the perspectives of degradation. To better record underwater scenarios, we have constructed and simulated 3-D underwater scenes using \textit{Blender} software. Its excellent ray tracing technology effectively replicates the propagation of light in water media, capturing physical phenomena such as scattering and absorption. Additionally, capturing the same 3-D scene in \textit{Blender} from different perspectives helps maintain consistency across viewpoints. Thus, this method is expected to not only simulate the degradation process of LF images more realistically but also ensure consistency across different views.

\section{Proposed Dataset}
\label{sec-dataset}
We build the first 4-D LF-based underwater imaging dataset with diverse degradations and rich scenes, namely LFUB. We use \textit{Blender}, an open-source 3-D modeling and rendering software, to simulate the effects of light propagation in a water medium to achieve as realistic underwater 4-D LF images as possible. This section outlines the construction pipeline of our dataset, including planning scene templates for building simulated environments, establishing degradation conditions that mimic real-world deterioration, and using camera arrays to render LF data. This approach not only serves our current project but also has potential applications in other LF tasks.

\subsection{Scene Construction}
To create highly realistic and dynamic simulations of underwater environments, we first compile a diverse collection of real underwater images sourced from Google and existing underwater image datasets \cite{li2019underwater}. These images serve as essential references for crafting our generated scenes. To further enhance the realism and variety of the scenarios, we download a selection of 3-D models from \textit{Sketchfab} \cite{ske}, carefully choosing those that closely align with objects typically found in underwater environments.
Using these models, we conscientiously construct underwater scenes that closely resemble those depicted in the templates. We strategically positioned the underwater objects within these scenes to authentically replicate natural underwater landscapes. This precise attention to object placement ensures that each scene closely resembles a genuine underwater environment, thereby improving the overall realism of our simulations.

In the initial stages of environment creation, we adopt the techniques recommended by the Infinigen \cite{raistrick2023infinite} to generate a realistic medium of water. This process involves creating a volumetric scattering shader to accurately mimic the subtle and complex light patterns characteristic of real underwater environments. However, challenges arise when using the generated water directly, as the inherent randomness in the simulation process limits control over specific outcomes. To overcome these limitations and introduce greater customization, we incorporate various color adjustments into the water body to simulate different types of water degradation. This step is essential for representing the diverse conditions found in different underwater environments, ranging from murky, particulate-laden waters to the clearer waters typical of open seas.

\begin{figure}[t]
\centering
\includegraphics[width=0.9\linewidth]{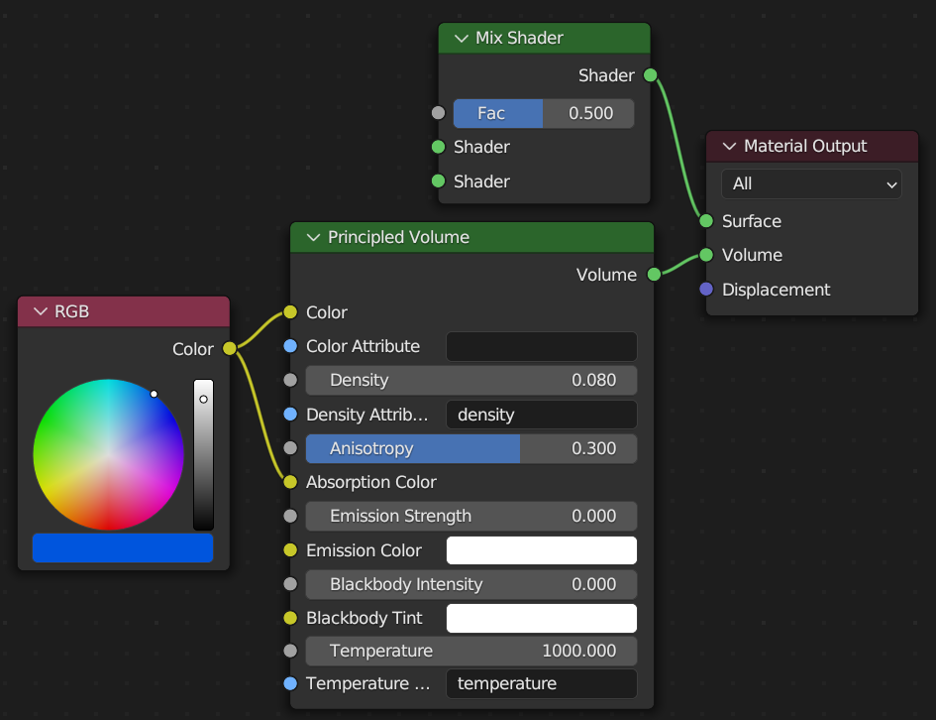}
\caption{\textit{Shader Editor} in \textit{Blender} software. In the \textit{Principled Volume} shader, we could choose different color deviations for underwater environments. The two main parameters \textbf{Anisotropy} and \textbf{Density} can be adjusted to change water effects.}
\label{fig:node}
\vspace{-3mm}
\end{figure}

\begin{figure}[t]
\centering
\subfloat[D=0.02]{%
  \includegraphics[width=0.23\linewidth]{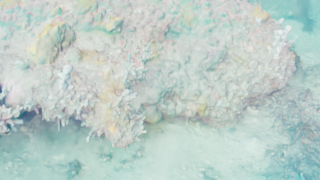}
}
\hfil
\subfloat[D=0.07]{%
  \includegraphics[width=0.235\linewidth]{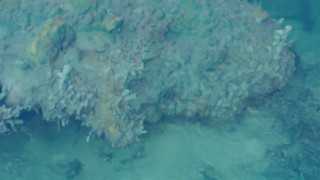}
}
\hfil
\subfloat[D=0.12]{%
  \includegraphics[width=0.235\linewidth]{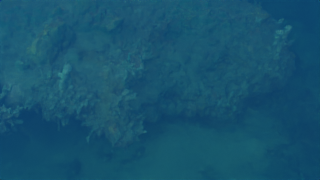}
}
\hfil
\subfloat[D=0.17]{%
  \includegraphics[width=0.235\linewidth]{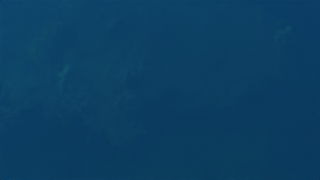}
}

\subfloat[A=0.0]{%
  \includegraphics[width=0.235\linewidth]{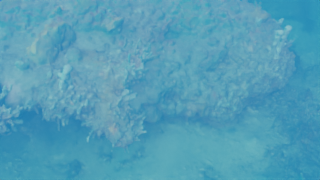}
}
\hfil
\subfloat[A=0.4]{%
  \includegraphics[width=0.235\linewidth]{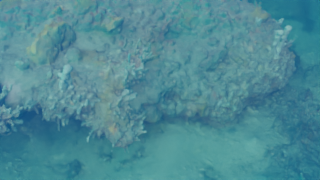}
}
\hfil
\subfloat[A=0.8]{%
  \includegraphics[width=0.235\linewidth]{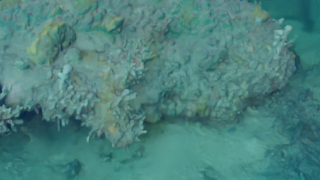}
}
\hfil
\subfloat[A=1.2]{%
  \includegraphics[width=0.235\linewidth]{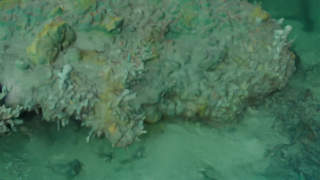}
}

\caption{Visual comparisons of different Anisotropy (A) and Density (D). Zoom in for details.}
\label{fig:para}

\end{figure}

While adjusting the visual properties of the water, we make extensive use of \textit{Blender}’s advanced shader capabilities, particularly focusing on the \textit{Principled Volume} shader, as shown in Fig. \ref{fig:node}. By manipulating the \textbf{Anisotropy} parameter, we can control the direction bias of specular reflections, allowing highlights to stretch along predominant directions. This adjustment closely mimics the way light behaves as it travels through water, which is often non-uniform and directionally biased due to factors like currents and particulate concentration. Additionally, we modify the \textbf{Density} parameter to alter opacity and the degree of light scattering within the water volume. Areas with higher density settings became more opaque, accurately reflecting the diverse visibility conditions commonly encountered in underwater environments. We show two groups of images with different Anisotropy and Density, in Fig. \ref{fig:para}. Specifically, we change the Density from 0.02 to 0.17 in the first line with the fixed Anisotropy. And we change the Anisotropy from 0 to 1.2 in the second line with the fixed Density. As we can see, the Density controls the lightness of the water and the Anisotropy controls the color of the water.

To complete our simulated underwater environments, we carefully arrange the collected 3-D objects within these scenes, ensuring that each setup accurately reflects varying depths, semantic contexts, and levels of environmental degradation. The crafted scenes range from shallow coral reefs brimming with marine life to deep-sea environments where light penetration is severely limited, each with distinct optical properties. By carefully tailoring the water's visual characteristics in each scenario, we achieve a diverse spectrum of underwater conditions. 

\subsection{Camera Setting}
In our pipeline of framing underwater scenes, we prioritize careful positioning of the camera to improve the relevance and quality of our training data. We strategically choose camera locations and angles to avoid capturing extensive areas of featureless water, which typically lack important visual information and might degrade the quality of the training dataset. To ensure accurate framing, we establish a shooting plane that is centered on the camera, with the cut plane oriented perpendicular to the camera's line of sight. This setup aids in maintaining a consistent orientation and depth of field across all captured images, thereby ensuring uniformity in the visual data collected. 
For rendering underwater 4-D LF images, we employ the \textit{Light Field Rendering Blender Plugin}\footnote{\href{https://github.com/gfxdisp/Blender-addon-light-field-camera/}{https://github.com/gfxdisp/Blender-addon-light-field-camera/}}, which enables the manipulation of various optical and geometric parameters to simulate realistic underwater conditions. We achieve array-style photography by translating the camera across the camera plane, where the translation distance $b$ represents the baseline of the LF, essentially the relative distance between adjacent cameras in the array. By adjusting this baseline, we can generate 4-D LF images with different ranges of parallax. This variability is crucial for training models to effectively perceive and interpret spatial relationships and depth in underwater environments.

\begin{figure}[t]
\centering
\subfloat[Types of color deviation degradation for two datasets. Left: 2-D RGB dataset UVBE \cite{xie2024uveb}. Right: 4-D LF dataset LFUB]{%
  \includegraphics[width=0.8\linewidth]{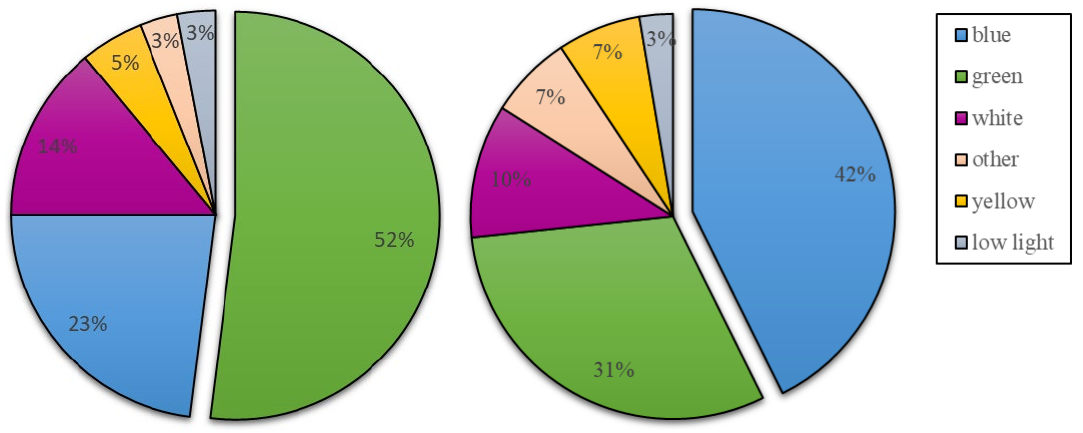}
  \label{fig:stat-degra}
}

\subfloat[Scene object categories]{%
  \includegraphics[width=0.8\linewidth]{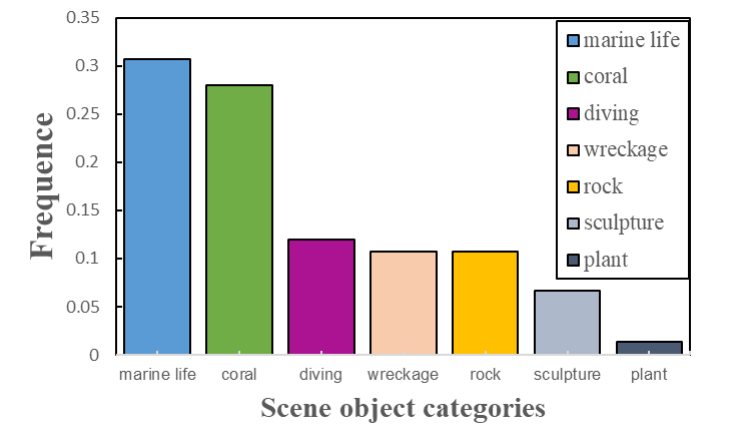}
  \label{fig:stat-object}
}

\caption{Statistics of the constructed underwater LF dataset. (a) Types of color deviation degradation for two datasets: the left one is 2-D RGB enhanced dataset UVBE \cite{xie2024uveb}, and the right one is our 4-D LFUB dataset. (b) Scene object categories.}
\label{fig:stat}
\vspace{-3mm}
\end{figure}

The disparity $d$ at different spatial locations in the LF is calculated using the formula:
\begin{equation}
    d = \frac{f \cdot b \cdot r}{s \cdot D},
\end{equation}
where $f$ is the focal length, $r$ is the output resolution, $s$ is the sensor size, and $D$ is the depth of scene location relative to the camera. By experimenting with different baselines $b$, we can produce a series of 4-D LF images, each offering a different perspective and depth, thereby enriching the dataset with diverse parallax ranges.

\subsection{Image Rendering}
After preparing the scene and setting the camera, we can directly render to obtain underwater LF images via \textit{Blender}. However, rendering LF images without the water medium is not feasible with the current setup, as the presence of water inherently affects the rendering process. To address this, in scenarios where water effects are not needed, we intentionally remove all water-related elements to produce corresponding scenes without water effects. This method allows us to preserve the original scene's characteristics while removing the influence of water, thus providing a direct comparison between underwater and no-water conditions, which is crucial for understanding the specific impact of water on the optical properties observed in the underwater scenes.

After rendering, the images undergo processing to set the point of infinite distance as the reference for zero parallax. To improve visual accuracy and meet particular analytical requirements, we apply a refocusing operation, as detailed by \cite{ng2006digital}, to shift the reference point to a designated area within the scene, thus re-calibrating the zero parallax point. This adjustment allows for a more precise analysis of parallax effects within a specific range, which in our dataset is set between $[-3,~3]$ pixels.

In summary, the conduction process of our dataset involves several key stages. First, we use real underwater photos as templates to guide the selection of 3-D models and the design of 3-D scenes. Next, we utilize \textit{Blender}'s shader system to generate degradation effects that closely replicate underwater conditions. In the well-designed 3-D scenes, cameras and camera planes are appropriately set up to frame suitable areas for rendering. The camera is then systematically translated across the camera plane to capture images from multiple perspectives. Finally, a refocusing technique is applied to all viewpoint images to adjust the focus.

\subsection{Data Format}
\begin{figure}[t]
\centering
\subfloat[]{%
  \includegraphics[width=0.45\linewidth]{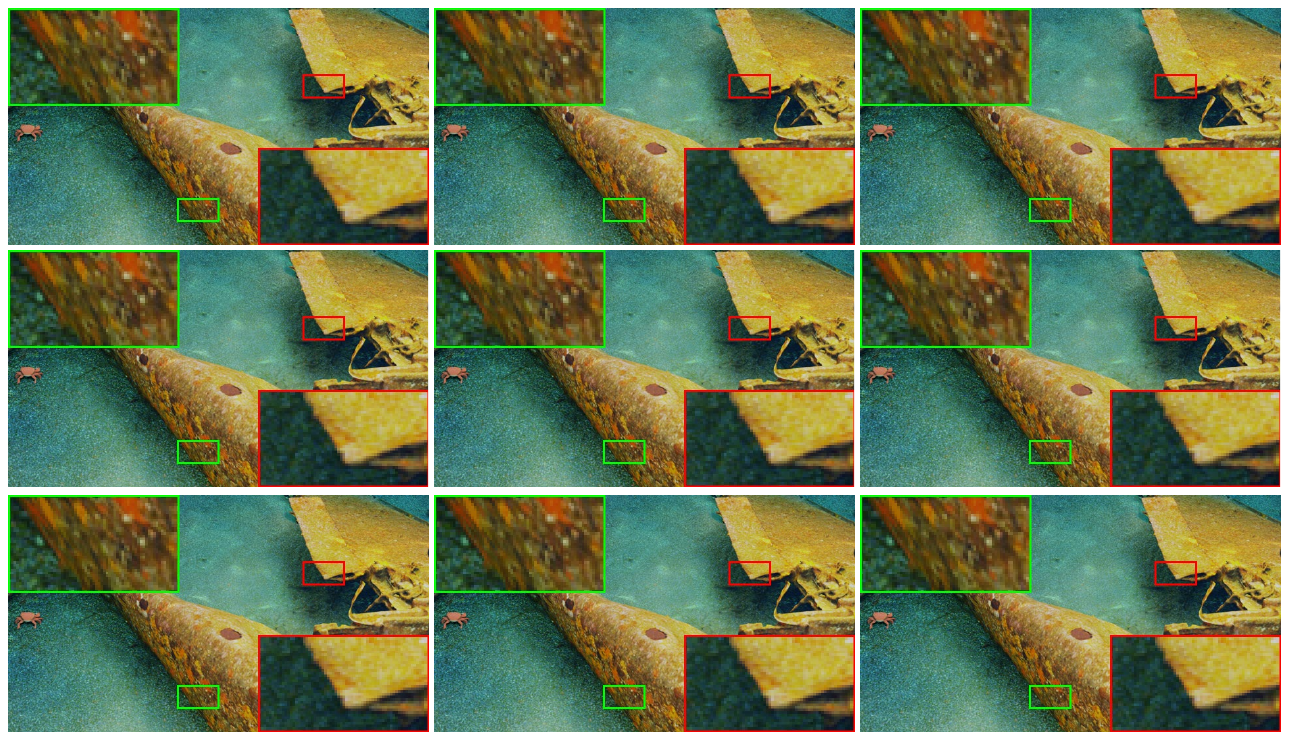}
  \label{fig:sai}
}
\subfloat[]{%
  \includegraphics[width=0.45\linewidth]{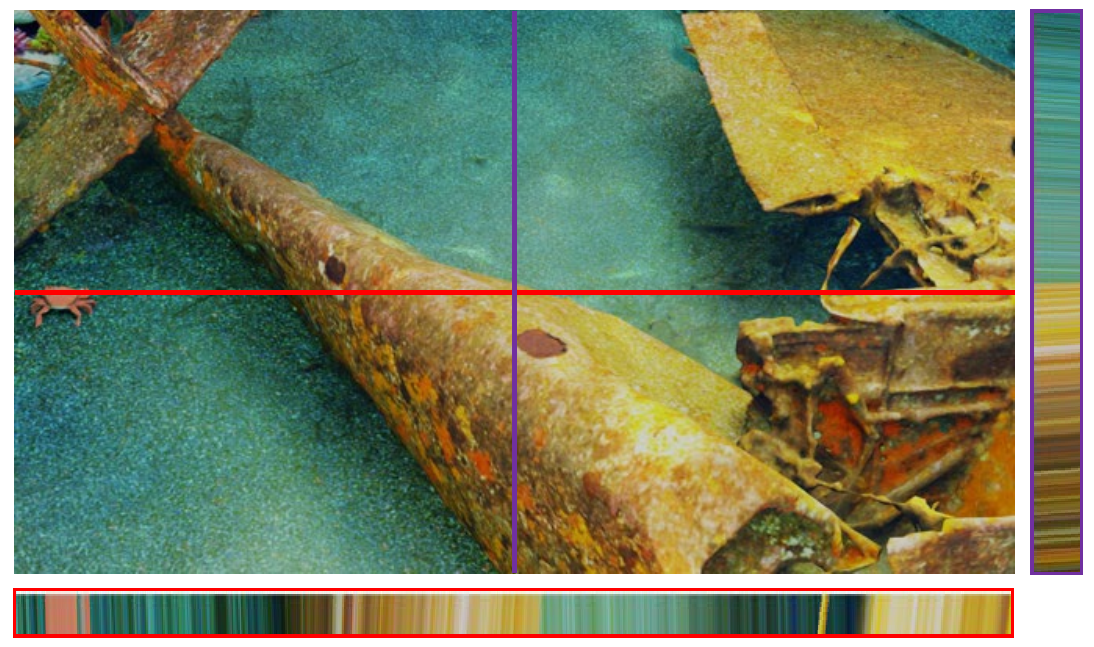}
  \label{fig:epi}
}
\caption{Two different forms of 4-D LF representation for one scene \textit{Plane}. (a) Sub-aperture images: the content of each magnified area varies with the view position; (b) Epipolar plane images: slices of 4-D LF. Zoom in for details.}
\label{fig:one-sample}
\vspace{-4mm}
\end{figure}

\textit{1) View Images.} Each scene has $7 \times 7$ views with each image of original 1920 $\times$ 1080 resolution. We have built 75 scenes. One scene sample is shown in Fig. \ref{fig:one-sample} with two different forms, including sub-aperture images (SAIs) and epipolar plane images, both of no-water condition. 

\textit{2) Depth Maps.} During the multi-view rendering process, we also export the ground truth depth maps of each scene. The samples of ground truth depth maps are provided in the bottom row of Fig. \ref{fig:sample-images}. 

\textit{3) Camera Parameters.} We provide the camera location, camera rotation, sensor size, and focal length of all view images. 

\textbf{Remark:} 1. The rendering process ensures that consistent degradation effects are applied to each individual view, making the LFUB dataset more realistic compared to previous synthetic LF-based low-light \cite{lamba2020harnessing} and rainy \cite{ding2021rain} datasets; 2. Single viewpoint images from the LFUB dataset can be used as a realistic evaluation set for traditional 2-D RGB-based enhancement algorithms \cite{li2021underwater, guo2023underwater}; 3. Notably, the LFUB dataset also provides valuable resources for multi-view learning tasks, such as scene reconstruction \cite{lyu2024rainyscape} and view synthesis \cite{guo2023content}; 4. The pipeline of conducting the LFUB dataset can be adapted to create other multi-view image datasets \cite{lyu2024enhancing, yan2022rain}; 5. The LFUB dataset is expected to facilitate the evaluation of LF disparity estimation methods \cite{wang2022disentangling, jin2022occlusion}.

\begin{figure}[t]
\centering
\includegraphics[width=1\linewidth]{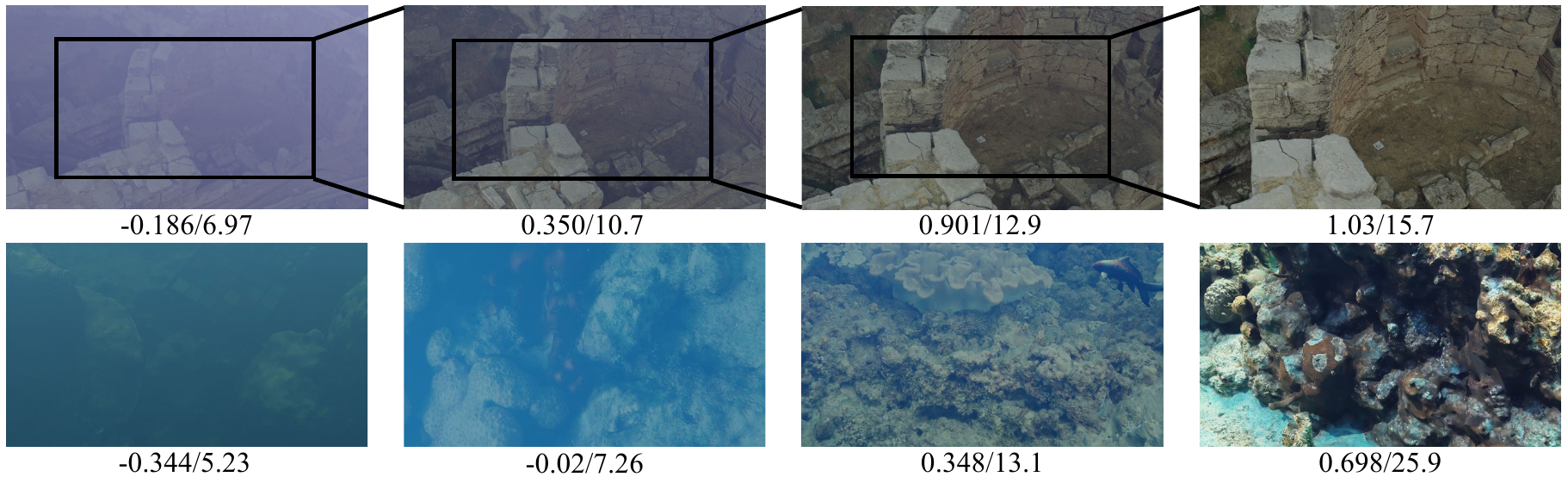}
\caption{The variation in the degree of degradation aligns with human perception, and underwater degradation indeed diminishes as depth approaches, consistent with physical processes in \textit{Blender}. The non-reference metrics UIQM and UCIQE increase as the visual quality improves. Zoom in for details.}
\label{fig:analysis}
\vspace{-4mm}
\end{figure}

\subsection{Statistics of LFUB Dataset}
\textit{1) Data Diversity.} Fig. \ref{fig:sample-images} shows the diversity of scenes and degradation types present in our LFUB dataset. To ensure a wide range of scenes, we utilize a variety of underwater 3-D models, as shown in Fig. \ref{fig:stat-object}, such as reefs (\emph{e.g.}, dead corals), marine animals (\emph{e.g.}, schools and turtles), and human activities underwater (\emph{e.g.}, divers and shipwrecks). To accurately represent real underwater conditions, we incorporate multiple types of degradation, including blue, green, yellow, other colors, and low-light conditions, as shown on the right side of Fig. \ref{fig:stat-degra}. We also provide the degradation statistics of UVEB \cite{xie2024uveb}, which contains real captured underwater images. The types of degradation in our LFUB dataset are similar to those of UVEB, with no significant differences in distribution, thereby affirming the validity of the degradation types included in our LFUB dataset.

\textit{2) Data Reliability.} \textit{Blender}'s built-in physically-based light transmission system allows for accurate simulation of underwater degradation effects. We calculate the non-reference underwater image quality metrics, including UIQM \cite{panetta2015human} and UCIQE \cite{yang2015underwater}. A higher UCIQE or UIQM score suggests a better human visual perception. As shown in Fig. \ref{fig:analysis}, we provide samples of different degradation with their image quality metrics. The first row shows that as the camera moves progressively closer to the scene, the severity of underwater degradation decreases, leading to improved metrics. This observation confirms that the underwater degradation modeled in \textit{Blender} diminishes with decreasing depth, in line with actual physical processes. The second row reveals that within images showing the same type of color deviation, the quality metrics increase as the degree of color distortion decreases. Thus, images with superior overall quality consistently achieve higher metric scores.

\begin{figure*}[!t]
\centering
\includegraphics[width=1\linewidth]{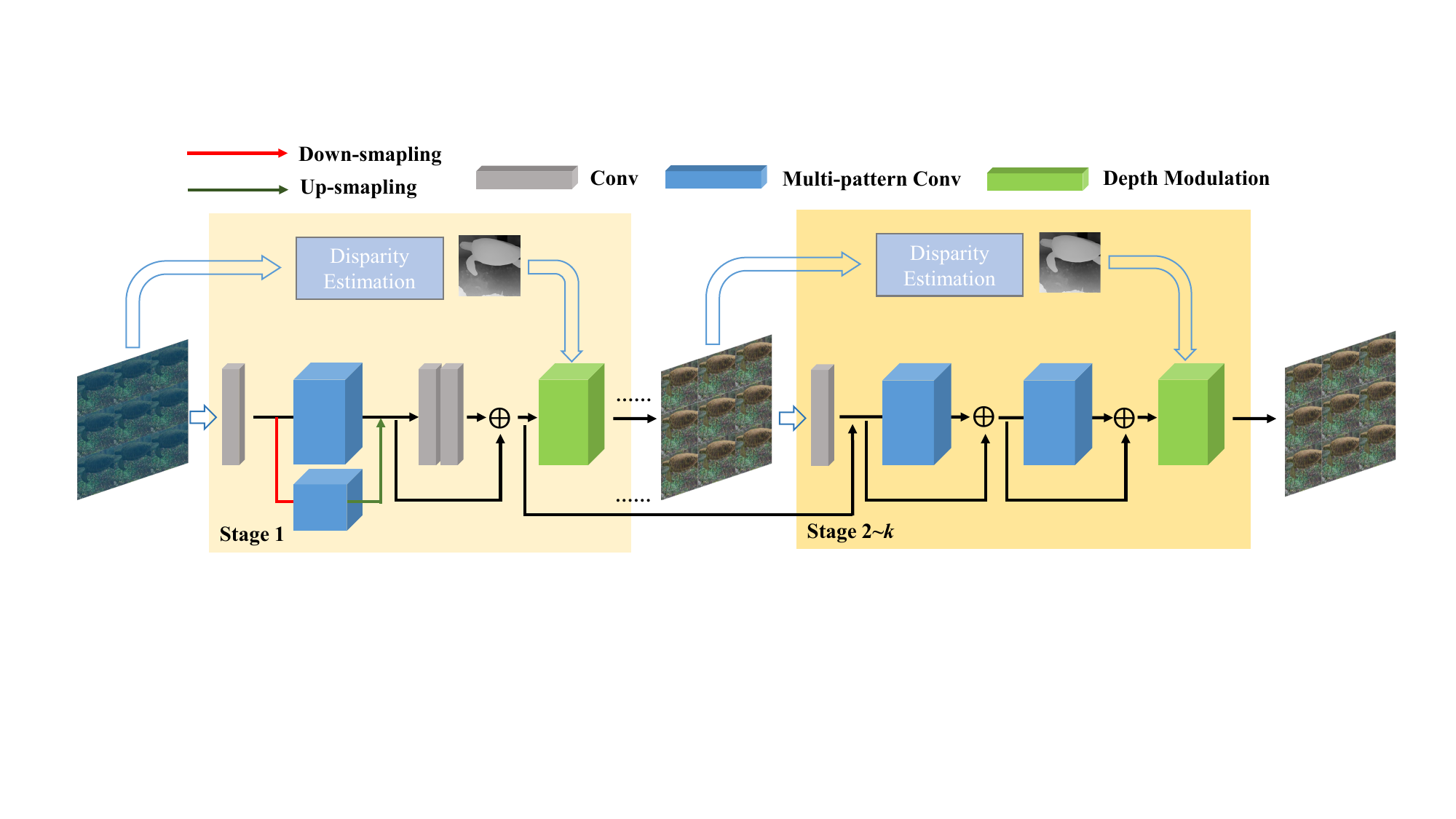}
\caption{Network architecture of our multi-stage network. In each stage, we estimate the central view disparity from the input of this stage with OccUnNet \cite{jin2022occlusion}. In the enhancement branch, we first extract features by multi-pattern convolution at different scales. Then, the multi-pattern features are passed to the Depth Modulation Block. We estimate dynamic kernels from EPI features, which are used to adjust spatial and angular features. At Last, the features are enhanced by the estimated disparities. The network then outputs the results from the enhanced features.}
\label{fig:pipeline}
\vspace{-4mm}
\end{figure*}

\section{Proposed Method}
\label{sec-method}

In this section, we propose a learning-based method for enhancing underwater 4-D LF images, as illustrated in Fig. \ref{fig:pipeline}. The framework follows the basic idea that depth estimation and the enhancement process depend on each other, and the network enhances underwater 4-D LF images through iterative stages. Specifically, in the initial stage, we feed the input to a multi-pattern feature interaction module (MFIM) at different scales. We then concatenate the multi-scale features passing to the depth module with the disparity from an estimator. In the depth modulation block, the former features are adjusted using explicit depth guidance and implicit EPI geometry convolution kernels. After modulation, the output features are passed through a convolution layer to reconstruct a result. In subsequent stages, the coarse result from the preceding stage is used as input, and the intermediate features from the previous stage are used for skip connection.

\subsection{Multi-pattern Feature Interaction Module}
The previously commonly used spatial-angular separable (SAS) convolution structure \cite{yeung2018light}, which includes spatial convolution and angular convolution, can only utilize local EPI information indirectly. This limitation hinders its ability to handle large disparity information or resist noise present in the LF data. Clearly, to fully address large disparity LF images, one should directly process EPIs. Therefore, similar to the method proposed by \cite{lyu2024enhancing}, we apply 2-D convolution to both spatial, angular, vertical EPI and horizontal EPI to achieve more thorough multi-dimensional feature interactions.

\begin{figure*}[h]
    \centering
    \begin{tikzpicture}
        \node[anchor=south west,inner sep=0] (image) at (0,0) {\includegraphics[width=0.9\textwidth]{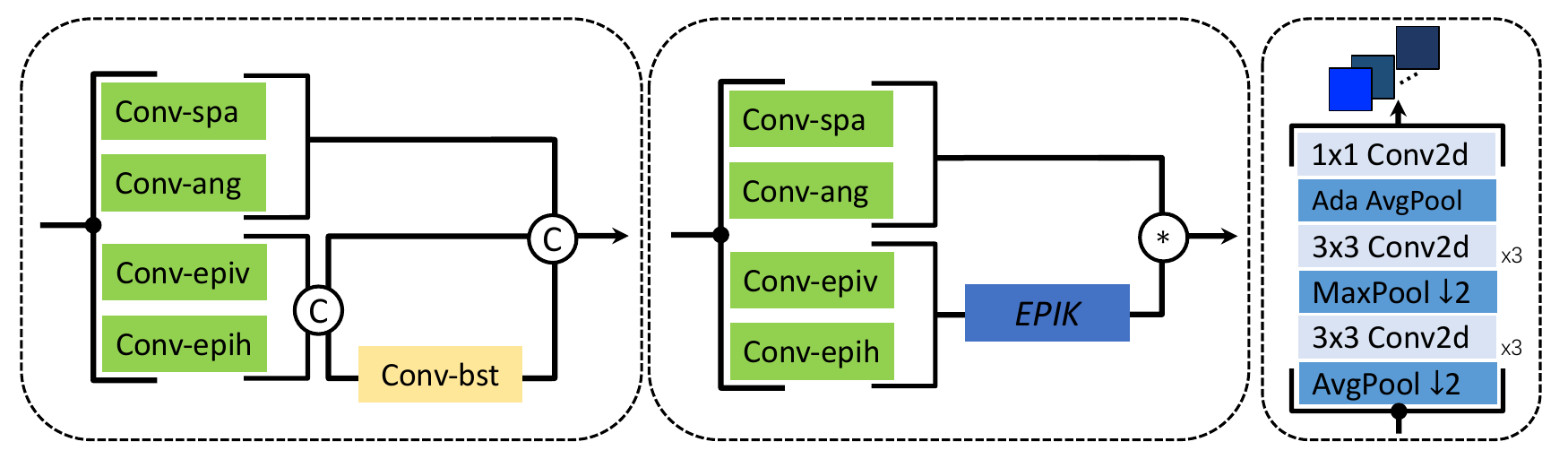}};
        \begin{scope}[x={(image.south east)},y={(image.north west)}]
            \node[align=center] at (0.215,-0.05) {(a) MFIM};
            \node[align=center] at (0.615,-0.05) {(b) EPI-IM};
            \node[align=center] at (0.9,-0.05) {(c) EPIK};
        \end{scope}
    \end{tikzpicture}
\caption{Network architecture of each component. (a) Multi-pattern Interaction Module (MFIM): each MFIM is composed of four-dimensional feature extraction and boosting convolution for EPI features; (b)EPI-guided implicit modulation (EPI-IM): two dimensions EPI features are used to generate convolution kernels (EPIK) for adjusting spatial and angular features; (c) EPIK: each EPIK is composed of several convolution layers and pooling layers.}
\label{fig:net-1}
\vspace{-4mm}
\end{figure*}

Initially, LF images are fed into a spatial feature extraction (SFE) module. We then use a multi-pattern feature interaction module to extract inter-view correlations. Following this, we apply 1 $\times$ 1 convolution to fuse different features, including four-dimensional features and boosted EPI features, as shown in Fig. \ref{fig:net-1}. Thus, the interaction process is defined as 
\begin{equation}
    \mathcal{F}=\mathrm{Conv}_{1\times1}([f_{spatial},f_{angular},f_{epiv},f_{epih}, f_{boost}]),
\end{equation}
where $f_{\cdot}$ denotes features extracted after SFE for different dimensions, i.e., spatial, angular, vertical EPI, horizontal EPI, and boosted EPI.

\subsection{Disparity/Depth Estimation}
The 4-D LF image implicitly encodes depth information by capturing a scene from multiple views, from which depth can be inferred by analyzing the differences between these perspectives. Previous LF image restoration methods typically explore the inter-view correlations that arise from disparities. Recently, several works \cite{jin2020deep, wang2022disentangling} have provided different disparity estimators for LF spatial SR and other tasks. However, it is difficult to obtain ground truth depth maps in practice. Furthermore, underwater imaging introduces additional challenges to depth estimation, such as scattering and absorption of light distorting the implicit depth cues. In response to these challenges, we utilize an occlusion-aware unsupervised LF image disparity estimation network named OccUnNet \cite{jin2022occlusion} to obtain the disparity map of the central view.

OccUnNet initially partitions a 4-D LF image into sub-LFs, each of which contains a certain subset of the LF subjected to 4-D transformations. These are subsequently processed through a multi-scale network that simultaneously produces the initial depth maps and their corresponding reliability maps. These maps then undergo back-transformations to align with the central SAI using an unsupervised constrained loss and a global smoothness loss. Ultimately, the final depth map is produced by an occlusion-aware fusion strategy. 

\subsection{Depth Modulation}
In general, the degradation process of underwater imaging can be expressed as \cite{tan2008visibility} 
\begin{equation}
    \mathbf{I} = \mathbf{T} \odot \mathbf{J} + (1-\mathbf{T})\odot \mathbf{A},\qquad \mathbf{T} = \mathrm{exp}(-\beta \cdot \mathbf{D}),
    \label{equ:degradationprocess}
\end{equation}
where $\mathbf{I}$ denotes the observed image; $\mathbf{J}$ is the clean image; $\mathbf{A}$ is the global background light; $\odot$ is the pixel-wise multiplication; $\mathbf{T}$ is the transmission map, which is directly related to depth $\mathbf{D}$ controlled by attenuation coefficient $\beta$. 

\begin{figure}[t]
\centering
\includegraphics[width=1\linewidth]{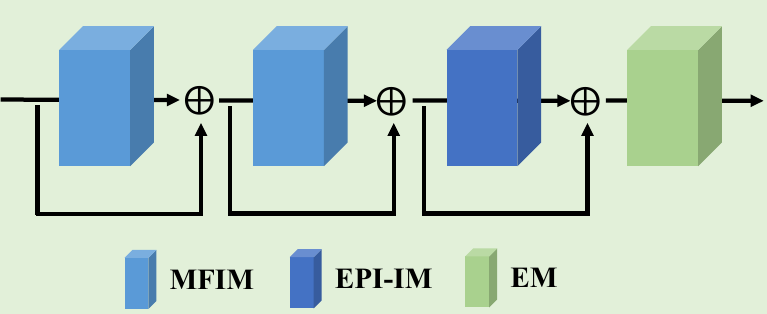}
\caption{The schematic illustration of the Depth Modulation module. Each depth modulation module is composed of two MFIM blocks, one EPI-IM, and one EM. DM block first estimates multi-pattern features and then generates dynamic kernels from EPI features to adjust spatial and angular features. At Last, the features are enhanced based on the estimated disparities.}
\label{fig:dm}
\vspace{-4mm}
\end{figure}

According to Eq. \eqref{equ:degradationprocess}, it is clear that depth plays a crucial role in influencing the extent of degradation across various regions. The 4D structure is equivalent to implicitly recording the depth information. Thus, we explore both explicit modulation (EM) and EPI-guided implicit modulation (EPI-IM) of geometry information for 4-D LF data, as illustrated in Fig. \ref{fig:dm}. Explicitly, we integrate depth into the network by passing the disparity of the central view through a 3 $\times$ 3 convolution layer, generating a deformation map $\boldsymbol{\theta}$ to modulate the features. Rather than utilizing depth solely for guiding feature extraction, we employ depth to direct the restoration at the end of each stage, aligning with the underwater degradation. The explicit modulation process can be formulated as
\begin{equation}
    \mathcal{F}_{explicit}=\boldsymbol{\theta} \odot \mathcal{F} + \mathcal{F},
\end{equation}
where $\mathcal{F}$ is features passed from the MFIM.

Furthermore, recognizing that EPIs indirectly encode geometric information \cite{zhou2023light}, we create convolutional kernels (EPIK) utilizing EPI features. Subsequently, spatial and angular features are adjusted based on geometric structural information, as depicted in Fig. \ref{fig:net-1}. The implicit adjustment process is formulated as 
\begin{equation}
    \mathcal{F}_{implicit}=\mathrm{EPIK}(f_{epiv}, f_{epih}) \ast [f_{spatial},f_{angular}] + \mathcal{F},
\end{equation}
where $\ast$ represents a 2-D convolution operation with kernels of $\mathrm{EPIK}(f_{epiv}, f_{epih})$. 

\subsection{Progressive Enhancement and Disparity Estimation}
Drawing upon the intuitive insight that superior 4-D LF images yield more precise depth maps, and vice versa, we adopt a cyclical approach between depth estimation and image refinement, employing a multi-stage scheme that fosters mutual advancement between depth estimation and image enhancement. 

In the earlier stage, we downsample to engage with features across various scales, mitigating issues related to noise and misalignment. Specifically, the SFE features are divided into different scales: original size and downsampled 2$\times$ with average pooling. These features are then channeled through diverse convolution dimensions to encourage interaction. Before outputting the enhanced outcomes, depth information is passed into the reconstruction block as guidance. In the later stages, a residual connection linking the prior and current stages assists in compensating for pixel data from adjacent pixels. Diverging from the first stage, subsequent stages receive inputs comprising coarse enhancement outputs and features predating the reconstruction in the preceding stage.

\subsection{Loss Function}
We train our framework by minimizing a composite loss function designed to enhance the accuracy and perceptual quality of LF image predictions. Our loss function is composed of several distinct terms, each targeting different aspects of the predicted output LF image $\tilde{\mathbf{I}}$ relative to the ground truth LF image $\hat{\mathbf{I}}$. The loss $\mathcal{L}$ can be denoted as
\begin{equation}
    \mathcal{L} = \mathcal{L}_{rec} + \mathcal{L}_{per} + \mathcal{L}_{ssim}.
\end{equation}

The first component, reconstruction loss, involves the L1 norm, which measures the absolute differences between the predicted LF images at various stages and the ground truth. Specifically, $\mathcal{L}_{rec}$ is expressed as
\begin{equation}
    \mathcal{L}_{rec} = \lambda_1 \left \| \tilde{\mathbf{I}} - \hat{\mathbf{I}}\right \|_1 + \sum_{i=1}^{k-1} w_i\left \| \tilde{\mathbf{I}}_i - \hat{\mathbf{I}}\right \|_1,
\end{equation}
where $\lambda_1$ and $w_i$, which is from an earlier stage, encourage fidelity at multiple levels of the network’s hierarchy.

We use perceptual loss \cite{johnson2016perceptual} as a higher-level assessment based on perceptual similarities between the final prediction and the ground truth. It evaluates how perceptually similar the predicted image is to the target, focusing on aspects like texture and structural integrity, and $\mathcal{L}_{per}$ is formulated as
\begin{equation}
    \mathcal{L}_{per} =\lambda_2 \mathrm{Per} \left( \tilde{\mathbf{I}} - \hat{\mathbf{I}}\right),
\end{equation}
where $\lambda_2$ is a constant hyper-parameter, $\mathrm{Per(\cdot)}$ is the perceptual loss. 

At last, we utilize the structural similarity index \cite{wang2004image} to measure the similarity between the predicted and true images in terms of color, contrast, and structure. The structural similarity loss $\mathcal{L}_{ssim}$ is denoted as
\begin{equation}
    \mathcal{L}_{ssim} =\lambda_3 \left( 1-\mathrm{SSIM}\left( \tilde{\mathbf{I}} - \hat{\mathbf{I}}\right)\right),
\end{equation}
where $\lambda_3$ is also a constant hyper-parameter, $\mathrm{SSIM(\cdot)}$ is the structural similarity index.

\begin{table*}[t]
\scriptsize
  \caption{Quantitative comparisons (PSNR/SSIM/inference time) of different methods on proposed LFUB dataset. The best second-best results are highlighted in \textbf{bold} and \underline{underline}. “↑" (resp. “↓") means the larger (resp. smaller), the better. All compared methods are retrained on our training set and evaluated on the testing set}
  \label{tab-results}
  \centering
  \small
  \setlength{\tabcolsep}{0.8mm}{
  \begin{tabular}{c|cccccc}
    \toprule[1.2pt]
    ~~~~Methods~~~~ 
    &~~~Fusion \cite{ancuti2012enhancing}~~~
    &~~~GDCP \cite{peng2018generalization}~~~
    &~~~MMLE \cite{zhang2022underwater}~~~ 
    &~~~WWPF \cite{zhang2023underwater}~~~ 
    &~~~LANet \cite{liu2022adaptive}~~~ 
    &~~~PUIE \cite{fu2022uncertainty}~~~ \\

    \midrule
    PSNR$\uparrow$    
    &14.19 &13.23 &13.92 &14.67 &17.97 &18.81   \\
    SSIM$\uparrow$    
    &0.3470 &0.4118 &0.4912 &0.5304 &0.7987 &0.7768         \\
    Inf. time$\downarrow$    
    &0.3512 &0.6469 &0.2894 &0.7557 &0.1122 &0.2823        \\ 
    % Memory(G)    
    % &- &- &- &- &2.73 &2.71        \\    
    
    \midrule[1.2pt]
     ~~~~Methods~~~~  
    &~~~Ushape \cite{peng2023u}~~~ 
    &~~~Uranker \cite{guo2023underwater}~~~
    &~~~UVE \cite{du2024end}~~~
    % &~~~UVENet \cite{xie2024uveb}~~~
    &~~~DistgSSR \cite{wang2022disentangling}~~~ 
    &~~~MSPNet \cite{wang2023multi}~~~ 
    &~~~Ours~~~ \\
    \midrule
    PSNR$\uparrow$    
    &18.29 &18.48  &\underline{19.91} &17.52 &18.92 &\textbf{22.51}       \\
    SSIM$\uparrow$    
    &0.7877 &0.8345 &\underline{0.8571} &0.7631 &0.8226 &\textbf{0.8680}       \\
    Inf. time$\downarrow$   
    &0.1836 &0.1727 &0.2449 &1.0269 &0.9196 &0.6987       \\ 
    % Memory(G)    
    % &11.20 &13.98 &- &30.27 &16.14 &25.17       \\   
    \bottomrule[1.2pt]
  \end{tabular}}
  \vspace{-4mm}
\end{table*}

\section{Experiments}
\label{sec-exp}
In this section, we begin by providing the implementation details, followed by an introduction to the experimental settings. Next, we compare our method against state-of-the-art methods and conduct a series of ablation studies to evaluate our framework. 

\subsection{Implementation Details}
We divided the proposed LFUB dataset into a training set containing 60 scenes and a test set with 15 scenes. Our LF images feature an angular resolution of 7 $\times$ 7 and an initial spatial resolution of 1920 $\times$ 1080, which is reduced to 640 $\times$ 360 for training purposes. During training, due to the memory constraints of the NVIDIA RTX 3090 GPU, LF images were randomly cropped into fixed patches of 128 $\times$ 128. The pixel values of all images were normalized to the range $[0,~1]$. To augment the data, we employed random cropping, rotation, and flipping operations. For testing, images were overlapped without additional data augmentation.

Our framework was developed by using Pytorch environment and Python 3.9.0, and was trained on an NVIDIA RTX 3090 GPU with 24GB of memory. We utilized Adam optimization for training, with hyperparameters $\beta_1=0.9$ and $\beta_2=0.999$. The learning rate was set at $0.0001$, and the batch size was kept at 1. We implemented a three-stage enhancement process. The constant hyperparameters for the loss function were defined as follows: $\lambda_1 = 1$, $\lambda_2 = 0.01$, $\lambda_3 = 0.01$, $w_1 = 0.05$, and $w_2 = 0.5$.

We found that disparity estimation with the constructed 4-D LF underwater data is challenging due to the rugged nature of the scenes in our LFUB dataset. To address this, we pre-trained OccUnNet on a synthetic LF dataset, HCI Dataset \cite{honauer2017dataset}, and then performed a warm-up operation on our 4-D LF underwater data by warping the side views to the central view.
We chose this network because it leverages LF parallax to deliver precise relative depth estimation, while its compact size enables efficient joint training with the main network.

\subsection{Experiment Settings}
\textit{1) Methods under comparison.}
We compared our method with various image/video/LF-based enhancement methods. These methods include four 2-D RGB physical model-based approaches, Fusion \cite{ancuti2012enhancing}, GDCP \cite{peng2018generalization}, MMLE \cite{zhang2022underwater}, WWPF \cite{zhang2023underwater}, and four 2-D RGB data-driven methods, LANet \cite{liu2022adaptive}, PUIE \cite{fu2022uncertainty}, Ushape \cite{peng2023u}, Uranker \cite{guo2023underwater}, and one video-based enhancement method, UVE \cite{du2024end}. As there are no existing methods specifically for enhancing 4-D LF-based underwater imaging, we also compared our framework with two other LF image enhancement methods, including DistgSSR \cite{wang2022disentangling} and MSPNet \cite{wang2023multi}. Our proposed framework is the first supervised learning-based method to enhance 4-D LF-based underwater imaging. For training and testing 2-D RGB-based methods, each SAI of LF images was treated as an individual sample. For the video-based method, we selected several SAIs adjacent in viewpoint to form a single input sequence. All comparative evaluations utilized the released codes of these methods to present results on our LFUB dataset.

\begin{figure*}[t]
\centering
\subfloat[Input]{%
  \includegraphics[width=0.108\linewidth,angle=90]{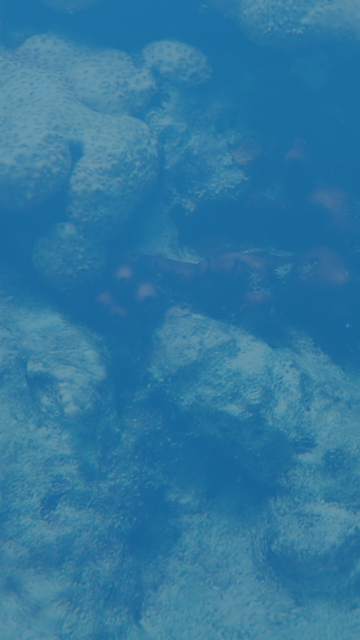}
}
\hfil
\subfloat[Fusion \cite{ancuti2012enhancing}]{%
  \includegraphics[width=0.108\linewidth,angle=90]{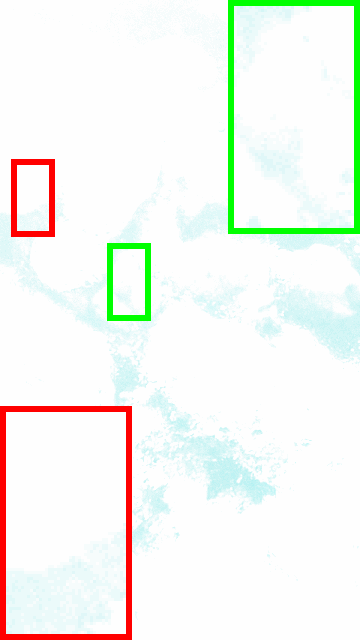}
}
\hfil
\subfloat[GDCP \cite{peng2018generalization}]{%
  \includegraphics[width=0.108\linewidth,angle=90]{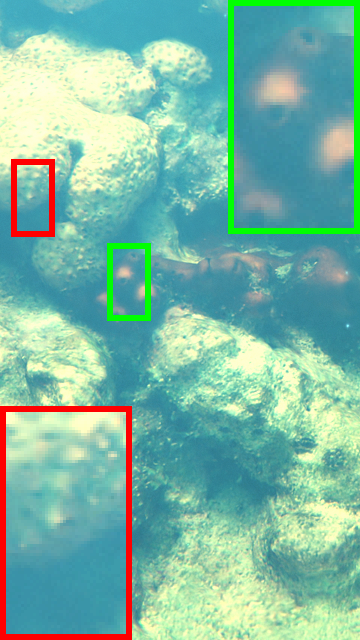}
}
\hfil
\subfloat[MMLE \cite{zhang2022underwater}]{%
  \includegraphics[width=0.108\linewidth,angle=90]{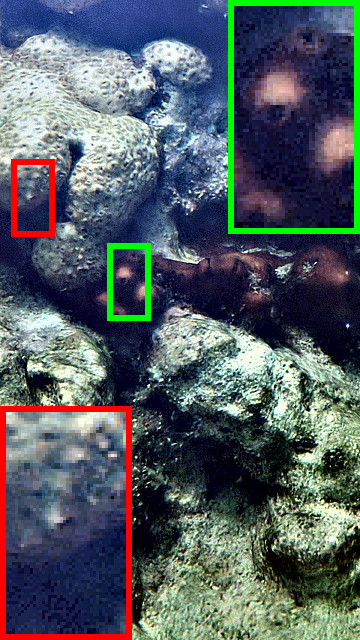}
}
\hfil
\subfloat[WWPF \cite{zhang2023underwater}]{%
  \includegraphics[width=0.108\linewidth,angle=90]{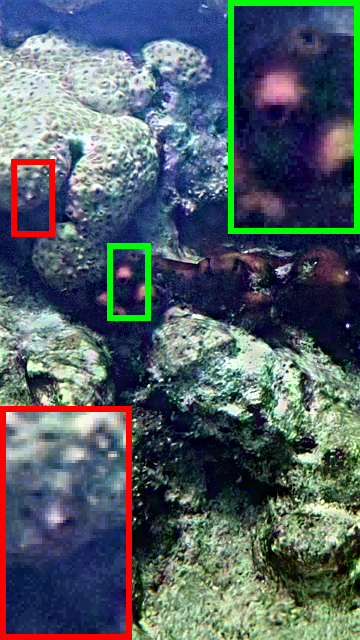}
}
\vspace{-0.3cm}

\subfloat[GT]{%
  \includegraphics[width=0.108\linewidth,angle=90]{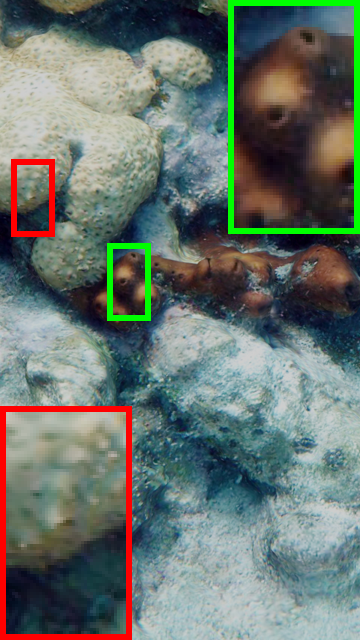}
}
\hfil
\subfloat[LANet \cite{liu2022adaptive}]{%
  \includegraphics[width=0.108\linewidth,angle=90]{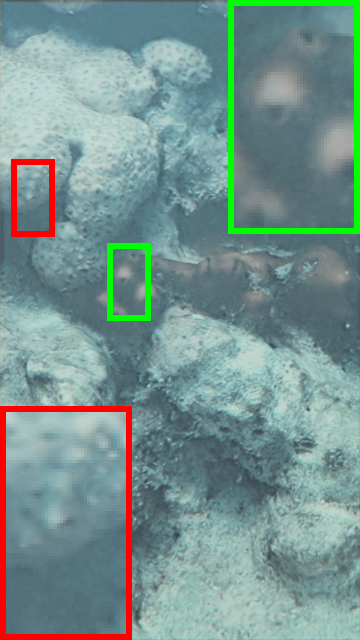}
}
\hfil
\subfloat[PUIE \cite{fu2022uncertainty}]{%
  \includegraphics[width=0.108\linewidth,angle=90]{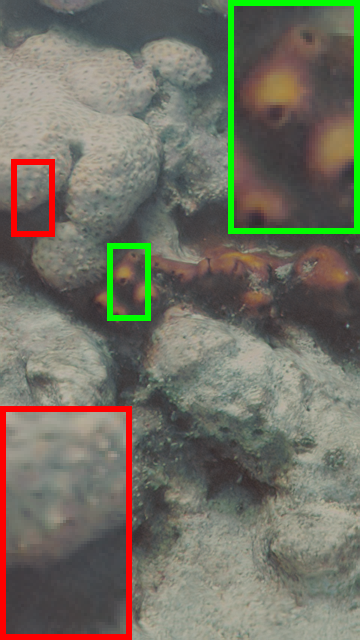}
}
\hfil
\subfloat[Ushape \cite{peng2023u}]{%
  \includegraphics[width=0.108\linewidth,angle=90]{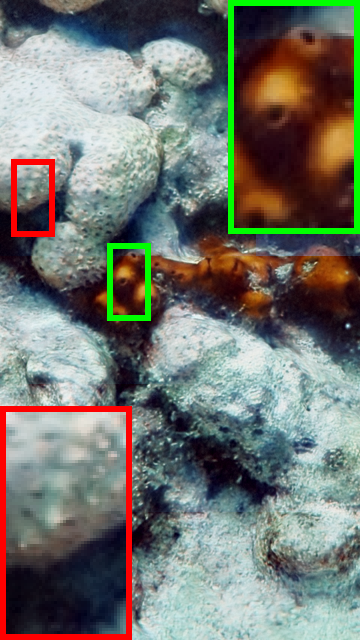}
}
\hfil
\subfloat[Uranker \cite{guo2023underwater}]{%
  \includegraphics[width=0.108\linewidth,angle=90]{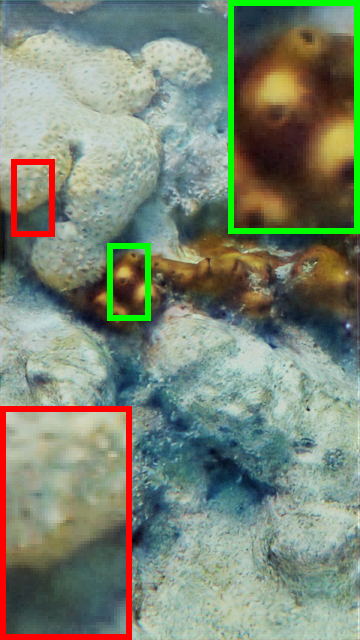}
}
\vspace{-0.3cm}

\hspace{0.1935\linewidth}
\subfloat[UVE \cite{du2024end}]{%
  \includegraphics[width=0.108\linewidth,angle=90]{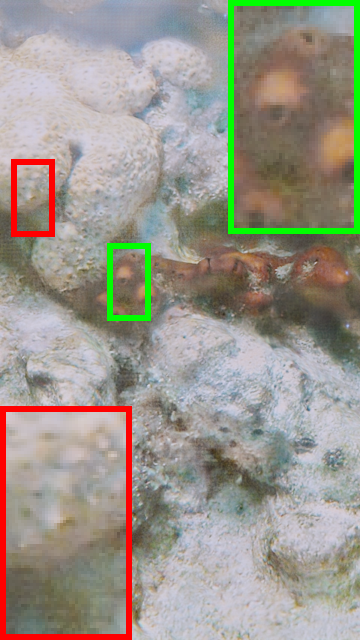}
}
\hfil
\subfloat[DistgSSR \cite{wang2022disentangling}]{%
  \includegraphics[width=0.108\linewidth,angle=90]{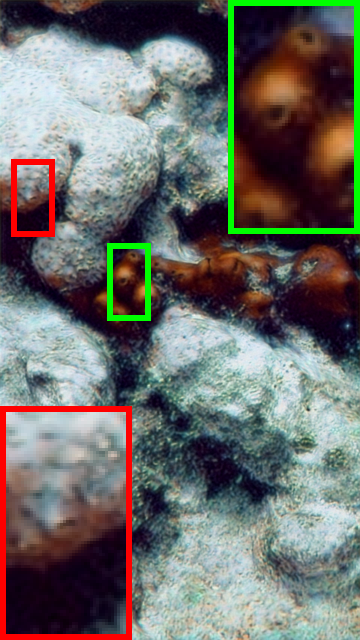}
}
\hfil
\subfloat[MSPNet \cite{wang2023multi}]{%
  \includegraphics[width=0.108\linewidth,angle=90]{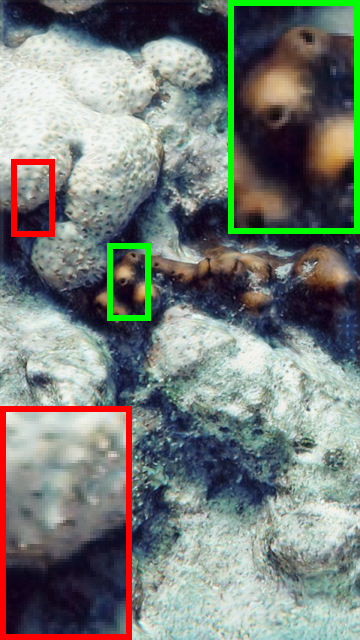}
}
\hfil
\subfloat[Ours]{%
  \includegraphics[width=0.108\linewidth,angle=90]{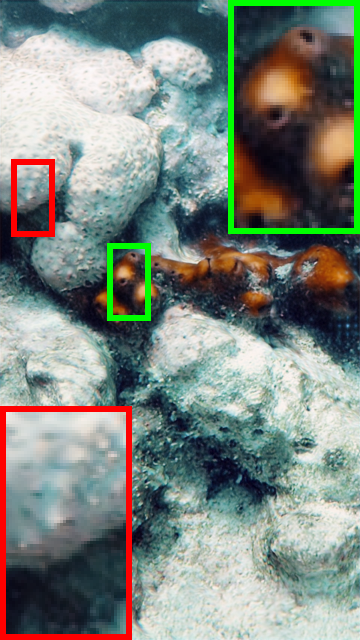}
}

\caption{Visual comparisons of different methods on \textit{Coral} with enlarged details.}
\label{fig:result-1}
\vspace{-4mm}
\end{figure*}

\begin{figure*}
\centering
\subfloat[Input]{%
  \includegraphics[width=0.192\linewidth]{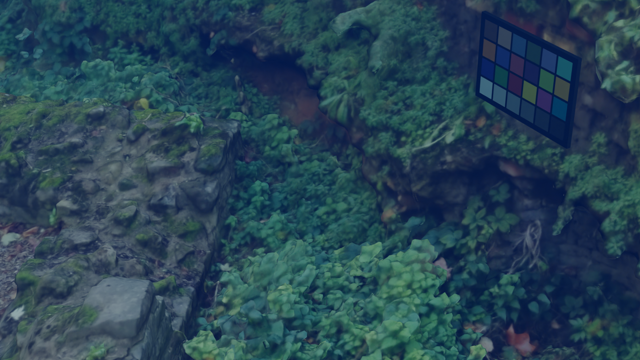}
}
\hfil
\subfloat[Fusion \cite{ancuti2012enhancing}]{%
  \includegraphics[width=0.192\linewidth]{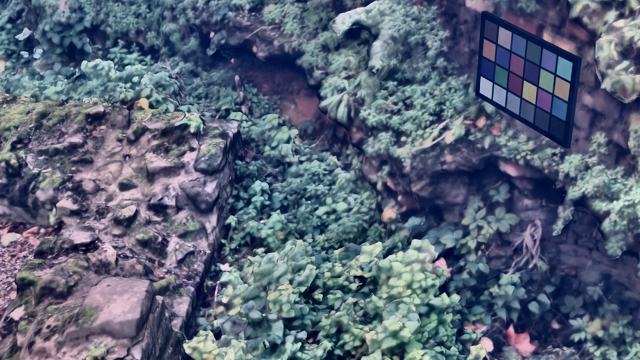}
}
\hfil
\subfloat[GDCP \cite{peng2018generalization}]{%
  \includegraphics[width=0.192\linewidth]{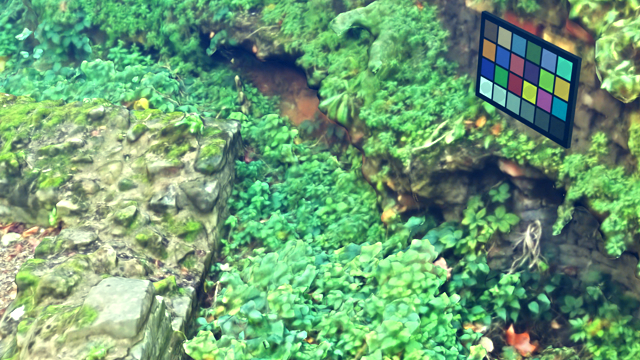}
}
\hfil
\subfloat[MMLE \cite{zhang2022underwater}]{%
  \includegraphics[width=0.192\linewidth]{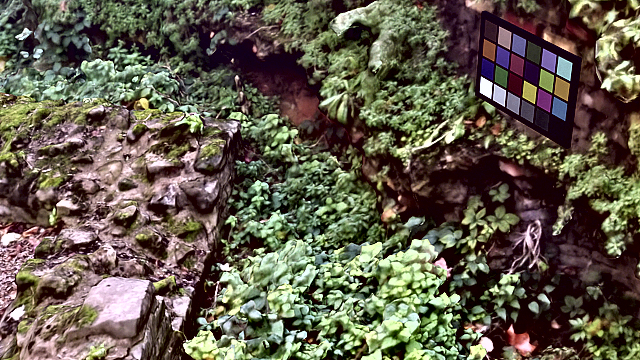}
}
\hfil
\subfloat[WWPF \cite{zhang2023underwater}]{%
  \includegraphics[width=0.192\linewidth]{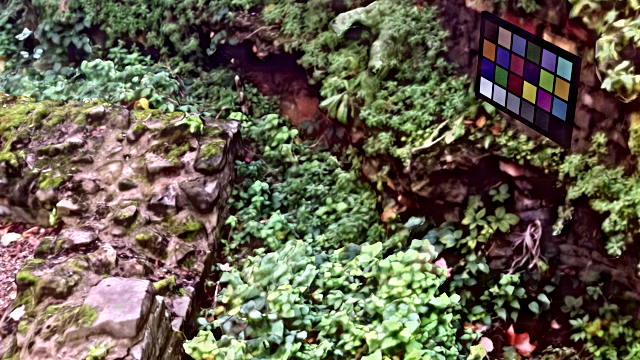}
}
\vspace{-0.3cm}

\subfloat[GT]{%
  \includegraphics[width=0.192\linewidth]{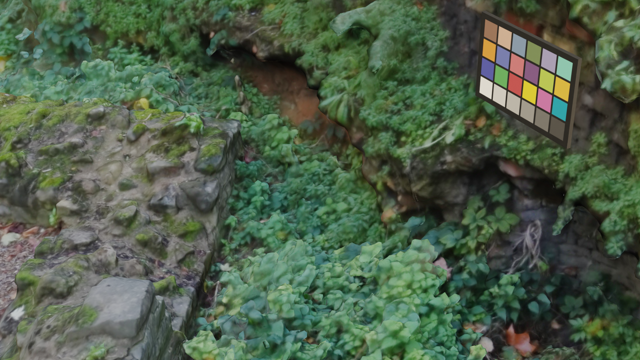}
}
\hfil
\subfloat[LANet \cite{liu2022adaptive}]{%
  \includegraphics[width=0.192\linewidth]{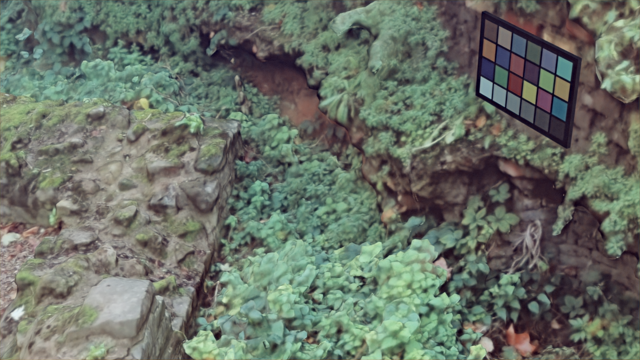}
}
\hfil
\subfloat[PUIE \cite{fu2022uncertainty}]{%
  \includegraphics[width=0.192\linewidth]{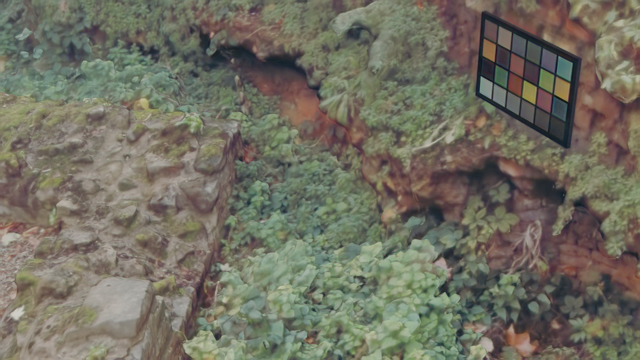}
}
\hfil
\subfloat[Ushape \cite{peng2023u}]{%
  \includegraphics[width=0.192\linewidth]{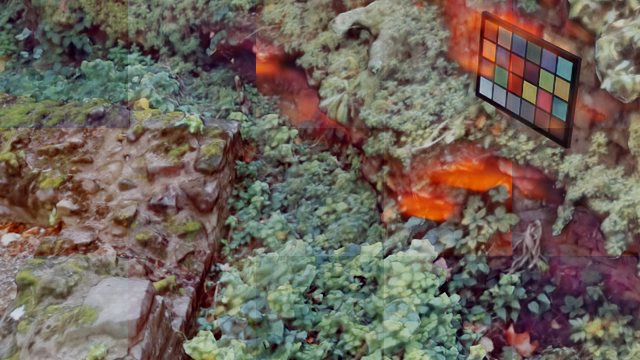}
}
\hfil
\subfloat[Uranker \cite{guo2023underwater}]{%
  \includegraphics[width=0.192\linewidth]{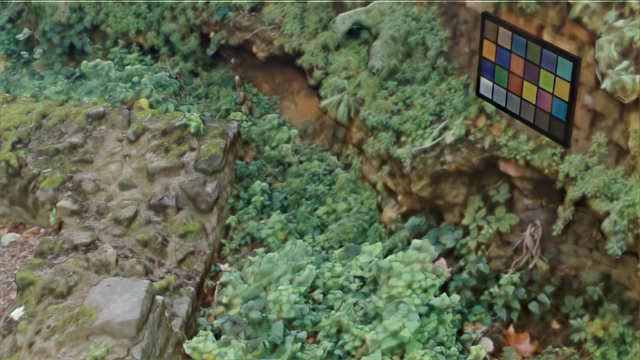}
}
\vspace{-0.3cm}

\hspace{0.1935\linewidth}
\subfloat[UVE \cite{du2024end}]{%
  \includegraphics[width=0.192\linewidth]{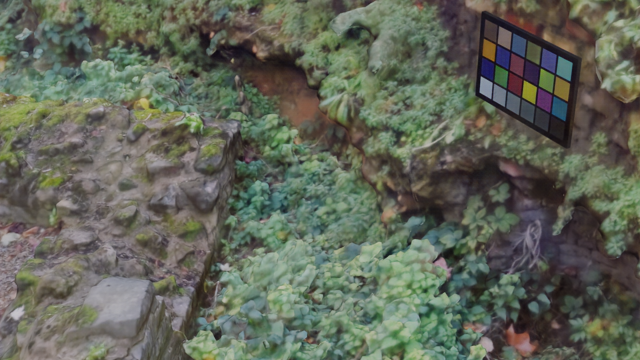}
}
\hfil
\subfloat[DistgSSR \cite{wang2022disentangling}]{%
  \includegraphics[width=0.192\linewidth]{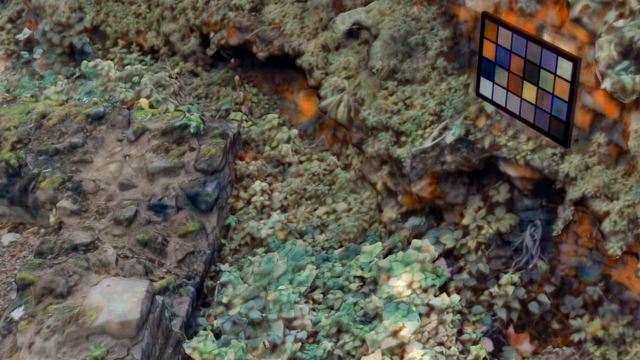}
}
\hfil
\subfloat[MSPNet \cite{wang2023multi}]{%
  \includegraphics[width=0.192\linewidth]{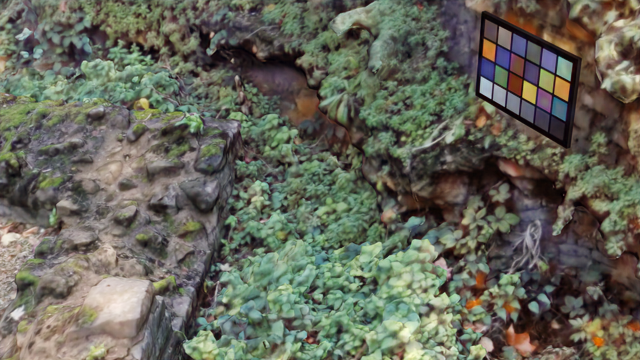}
}
\hfil
\subfloat[Ours]{%
  \includegraphics[width=0.192\linewidth]{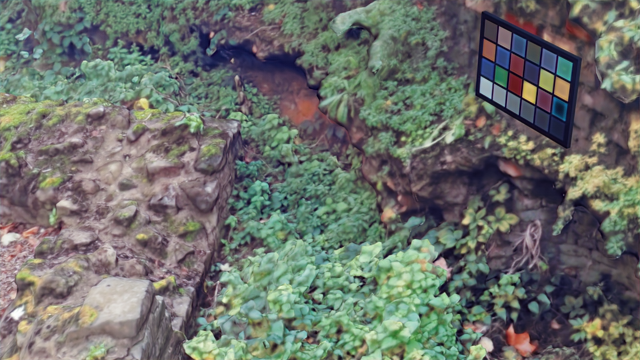}
}

\caption{Visual comparisons of different methods on \textit{Rock}.}
\label{fig:result-2}
\vspace{-4mm}
\end{figure*}

\textit{2) Evaluation Metrics.}
We employed widely-used metrics to quantitatively evaluate the performance of our method and compared methods. Peak Signal-to-Noise Ratio (PSNR) measures the image quality and underwater scene enhancement performance, with higher values indicating better results. Structural Similarity Index Measure (SSIM) assesses quality via structural information, luminance, and contrast, again with higher values denoting better outcomes. Additionally, we enlisted volunteers to rate the visual quality of enhancement results from different methods.

\begin{figure*}[t]
    \centering
    \begin{tikzpicture}
        \node[anchor=south west,inner sep=0] (image) at (0,0) {\includegraphics[width=0.9\textwidth]{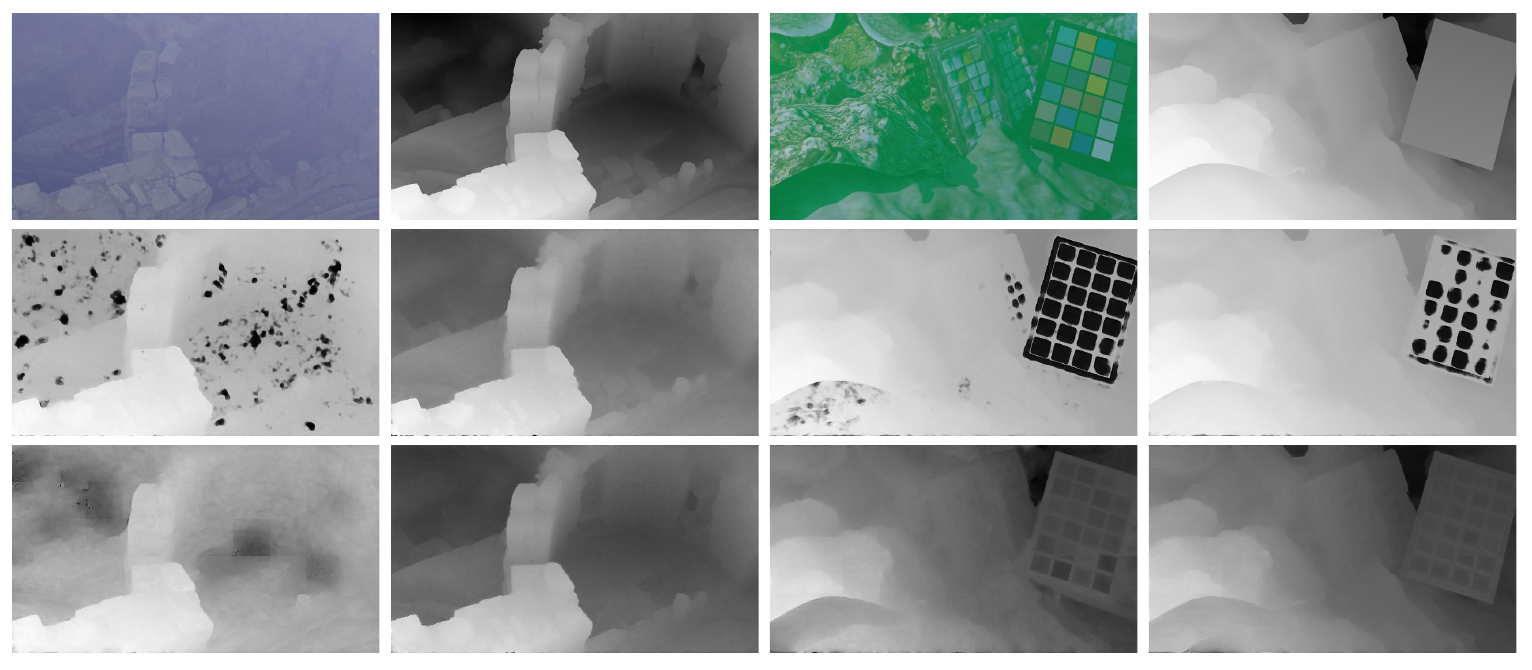}};
        \node[align=center, rotate=90] at (-0.1,5.9) {Input};
        \node[align=center, rotate=90, font=\small] at (-0.1,3.6) {DistgDisp \cite{wang2022disentangling}};
        \node[align=center, rotate=90, font=\small] at (-0.1,1.28) {OccUnNet \cite{jin2022occlusion}};
        \node[align=center, font=\small] at (2.1,7.2) {\textit{Statue}};
        \node[align=center, font=\small] at (6.1,7.2) {Depth GT};
        \node[align=center, font=\small] at (10.1,7.2) {\textit{Rock}};
        \node[align=center, font=\small] at (14.2,7.2) {Depth GT};
        \node[align=center, font=\small] at (2.1,-0.05) {Depth for Input};
        \node[align=center, font=\small] at (6.1,-0.05) {Depth for Result};
        \node[align=center, font=\small] at (10.1,-0.05) {Depth for Input};
        \node[align=center, font=\small] at (14.2,-0.05) {Depth for Result};
    \end{tikzpicture}
    \vspace{-2mm}
    \caption{Disparity maps estimated by methods DistgDisp \cite{wang2022disentangling} and OccUnNet \cite{jin2022occlusion} on proposed LFUB dataset. Top row: (from left to right) degraded \textit{Statue} image, \textit{Statue} GT Depth map, degraded \textit{Rock} image, \textit{Rock} GT Depth map. Middle row: disparity maps estimated by DistgDisp of the degraded and restored scenes. Bottom row: disparity maps estimated by OccUnNet of the degraded and restored scenes.} 
    \label{fig:depth}
    \vspace{-4mm}
\end{figure*}

\begin{figure}[t]
\centering
\includegraphics[width=0.9\linewidth]{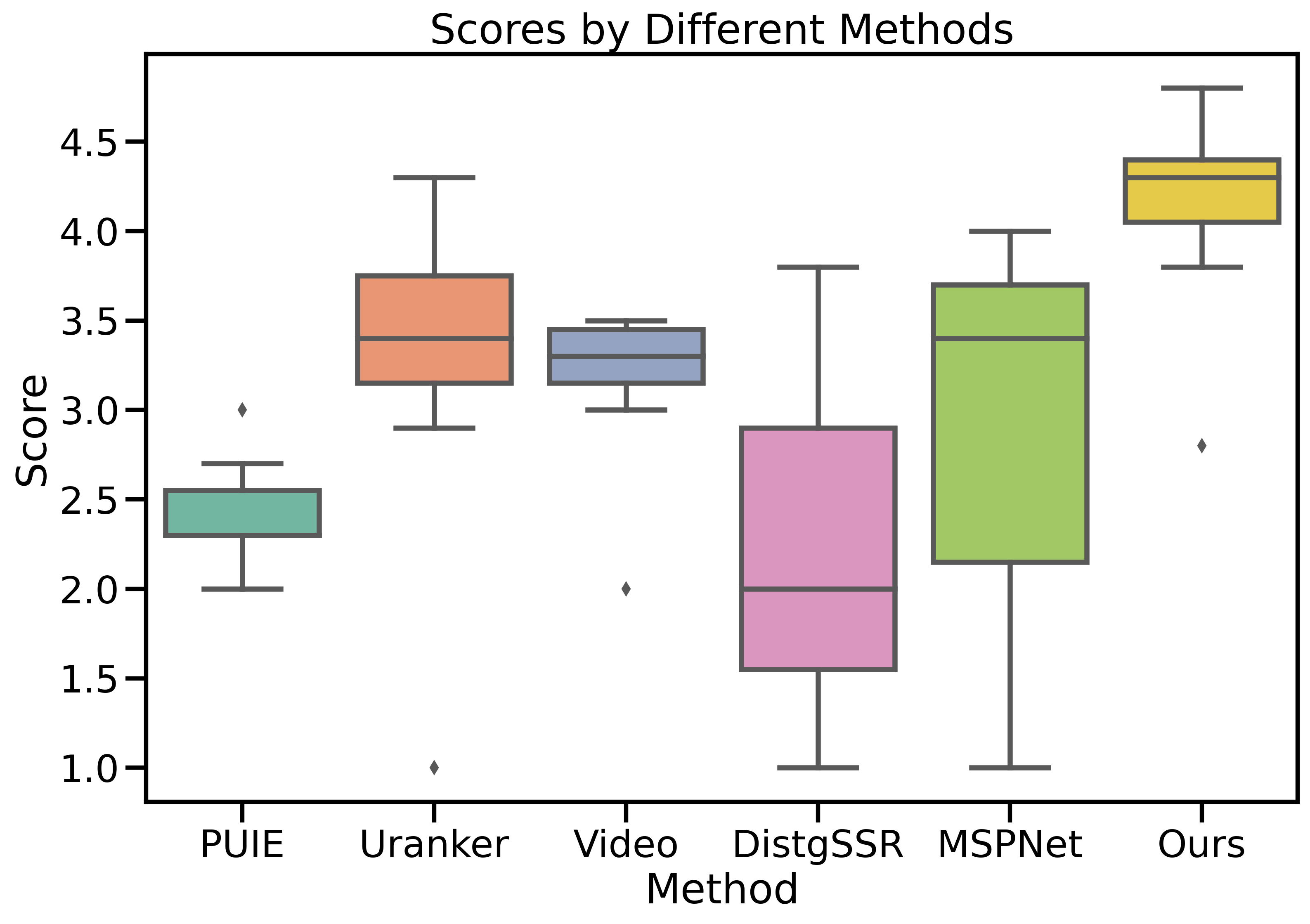}
\caption{User study results. Boxplot analysis illustrating the distribution of user study scores for different methods on the LFUB test set. Scores range from 1 to 5, with higher scores indicative of greater user recognition of image quality.} 
\label{fig:us}
\vspace{-4mm}
\end{figure}

\subsection{Network Architecture Evaluation}
We summarize the statistical results in Table \ref{tab-results}, showing quantitative results of all compared methods on our test set. Besides, Fig. \ref{fig:result-1} and Fig. \ref{fig:result-2} provide a visual comparison of the enhancement results from different methods. 

The scores of PSNR and SSIM metrics of different methods are reported in Table \ref{tab-results}. Our proposed method demonstrates the best performance as presented. Unlike 2-D RGB-based methods, the geometric information from the 4-D LFs proves advantageous in improving underwater imaging quality. Our specific strategy enables our framework to outperform two other LF-based methods, specifically due to our meticulous design for underwater environments. As for reference time, three LF methods have the capability to enhance all views of a scene simultaneously, whereas other methods require processing each image individually.

We initiate our analysis with the scene \textit{Coral}, which exhibits a pronounced bluish color deviation, as depicted in Fig. \ref{fig:result-1}. In Fig. \ref{fig:result-1}(a), this deviation significantly conceals structural details within the underwater environment. Regarding color correction, only DistgSSR \cite{wang2022disentangling}, MSPNet \cite{wang2023multi}, and our method successfully eliminates the bluish tint without introducing additional color artifacts. Notably, our method excels in restoring colors in coral and rock regions, as highlighted in the magnified areas of Fig. \ref{fig:result-1}(l)-(n), illustrating the effectiveness of our uniquely designed interactive structure for enhancing underwater imaging. Further, in Fig. \ref{fig:result-2}, we present visual results from different methods applied to the scene \textit{Rock}. Our method outperforms others, correcting the color checker with superior contrast, whereas other methods yield lower color intensity.

We advance our research through a detailed user study involving 14 participants to quantitatively validate the quality of image enhancement. For this analysis, we randomly choose seven scenes exhibiting varying degrees of color deviation from our test set. Participants evaluate the enhanced results against a set of criteria including color fidelity, noise, and structural integrity, using a 5-point Likert scale, with 1 signifying ‘bad’, 2: ‘poor’, 3: ‘fair’, 4: ‘good’, and 5: ‘excellent’. The evaluative criteria cover several aspects, including brightness, color fidelity, noise, and structural integrity. As depicted in Fig. \ref{fig:us}, the results of this study strongly support the superiority of our method over other competing techniques, substantiated by a thorough statistical analysis of the aggregated scores.

\subsection{Ablation Study}
We conduct a series of ablation studies to better understand the components of our method. We use 5 $\times$ 5 LF images in the following ablation experiments.

\textit{1) Evaluation of Different Configurations.}
Table \ref{tab-abfeature} shows the ablation studies of the components of our proposed network, including the multiple feature interaction module, explicit modulation, and EPI-guided implicit modulation. To conclude, the entire network achieves better enhancement performance than other ablated models on our test set, which suggests the effectiveness of the combinations of all components.
\begin{table}[t]
\centering
\caption{Quantitative results of ablation studies assessing the impact of various network architecture components within our proposed network. These components include network modules such as Multi-pattern Feature Interaction Module (MFIM), Explicit Modulation (Explicit), and EPI-guided Implicit Modulation (Implicit). ``\checkmark'' (resp. blank) represents the corresponding structure is used (resp. unused)}
\label{tab-abfeature}
\setlength{\tabcolsep}{3mm}{
\small
\begin{tabular}{c|cccc}
\toprule[1.2pt]
 & (a) & (b) & (c) & (d) \\ 
\midrule
MFIM     & \checkmark & \checkmark & \checkmark & \checkmark \\ 
EM &            & \checkmark &            & \checkmark \\ 
EPI-IM &            &            & \checkmark & \checkmark \\ 
\midrule
PSNR $\uparrow$    & 20.57  & 20.78  & 21.49  & \textbf{21.82} \\ 
SSIM $\uparrow$    & 0.8432 & 0.8491 & 0.8610 & \textbf{0.8641} \\ 
Inf. time (s)      & 0.0834 & 0.9109 & 0.0934 & 1.1253  \\ 
Memory (G)         & 19.41  & 20.08  & 21.98  & 22.12 \\ 
\bottomrule[1.2pt]
\end{tabular}}
\vspace{-4mm}
\end{table}

\begin{table}[t]
\caption{Quantitative results of ablation study about different depth information sources, including DistgDisp, OccUnNet, and Ground Truth.}
\label{depth-exp}
\centering
\small
\begin{tabular}{c|cccc}
\toprule[1.2pt]
Stages & PSNR $\uparrow$  & SSIM $\uparrow$ \\ 
\midrule
DistgDisp        & 21.15 & 0.8474    \\
OccUnNet         & 21.82 & 0.8641     \\
Ground Truth     & 22.21 & 0.8701    \\
\bottomrule[1.2pt]
\end{tabular}
\end{table}

\begin{table}[t]
\caption{Quantitative results of ablation study about different numbers of iterative stages. Numbers 1 to 5 represent the iterative stages}
\label{tab-abstage}
\centering
\small
\begin{tabular}{c|cccc}
\toprule[1.2pt]
Stages & PSNR $\uparrow$  & SSIM $\uparrow$ & Inf. time (s) & Memory(G) \\ 
\midrule
Ours-1     & 21.15 & 0.8474 &0.0818 & 22.03    \\
Ours-2     & 21.63 & 0.8570 &0.0912 & 22.08    \\
Ours-3     & 21.82 & \textbf{0.8641} &1.1253 & 22.12    \\
Ours-4     & \textbf{21.83} & 0.8636 &1.9258 & 22.17    \\ 
Ours-5     & 22.02 & 0.8637 &3.0448 & 22.21    \\ 
\bottomrule[1.2pt]
\end{tabular}
\end{table}

\begin{figure}[t]
\centering
\includegraphics[width=0.9\linewidth]{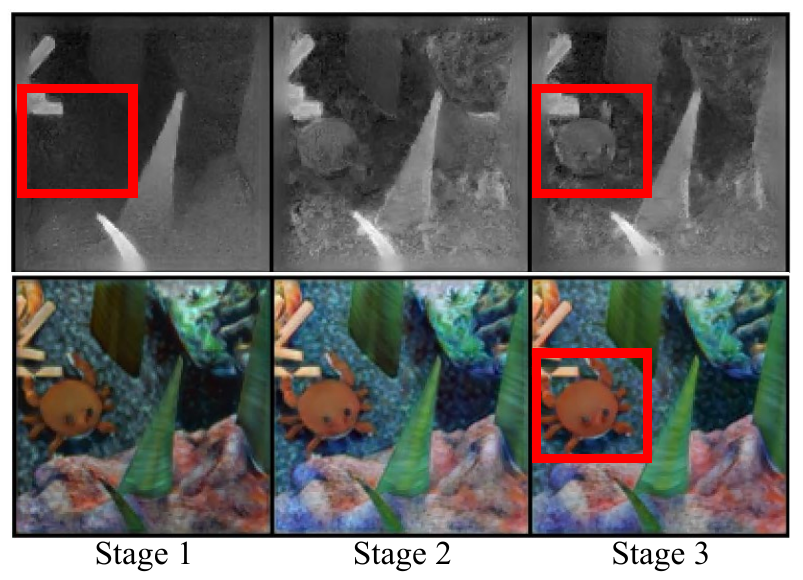}
\caption{Visual intermediate results of stages of (1,2,3). The first row is the estimated disparities of each stage; the second row is the corresponding enhanced results.}
\label{fig:itera}
\vspace{-4mm}
\end{figure}

\textit{2) Evaluation of Depth Estimation.}
Severe underwater degradation can compromise the content of images, leading to poor depth maps. As illustrated in the first and third columns of Fig. \ref{fig:depth}, the depth maps derived from estimated underwater images are subpar and lack a clear structural hierarchy. Conversely, as shown in the second and fourth columns of Fig. \ref{fig:depth}, the depth maps estimated from GT and restored image demonstrate superior results, providing sharp edges and preserving smoothness in areas of uniform depth. These results highlight the effectiveness of our framework with OccUnNet in providing better depth maps of enhanced LF images. And we compare different depth information, including ground truth and estimations from DistgDisp and OccUnNet, in Table \ref{depth-exp}. The quantitative results show that OccUnNet can achieve better performance than DistgDisp, but worse than the ground truth.

\textit{3) Evaluation of Iterative Stages.}
Fig. \ref{fig:itera} shows the central view of the enhanced LF image and the corresponding estimated disparity map after each stage. 
With the process of iteration between enhancement and disparity estimation, the hierarchy of the depth map becomes distinct. As color correction improves through different stages, the depth of the crab marked with a red box in the depth map is also gradually estimated.

We further conduct experiments on our proposed network with different stages (1, 2, 3, 4, 5). We train our model on our dataset with 5 × 5 angular resolution. Table \ref{tab-abstage} shows the results of different stages. 
Overall, adding more processing stages can enhance performance. However, beyond four stages, the improvement in enhancement performance becomes less significant. Additionally, more stages mean more parameters, which can slow down the model running. Therefore, we have made a trade-off between the model's running speed and performance, ultimately setting the stages as 3.

\subsection{Limitation and Failure Cases}

The currently designed algorithm does not take into account the methods
for deep-sea low-light and strong noise conditions. This situation will affect the depth estimation effect, and affect the interaction of information between different perspectives, affecting the actual restoration effect. When applied to areas with diverse color changes, our method cannot restore all colors. We show one case in the supplementary materials.

\section{Conclusion and Discussion}
\label{sce-c}

In this paper, we have made the first attempt to explore the acquisition of high-quality images in underwater environments through 4-D LF-based imaging. To this end, we have constructed a varied and photorealistic 4-D LF-based underwater image dataset, which serves as a benchmark for future research in this area. Additionally, we presented a novel approach to enhance underwater 4-D LF images, incorporating both explicit and implicit use of depth information from 4-D LFs. Our method has demonstrated significant improvements in color correction and overall image quality through the progressive refinement of depth estimation and image enhancement. Extensive experiments have validated that our approach achieves state-of-the-art performance in enhancing 4-D LF-based underwater imaging, effectively addressing the challenges posed by light absorption and scattering in underwater environments. 

In the future, the following directions could be considered for improving the proposed paradigm. We first plan to collect real underwater LF images to fulfill the need for performance evaluation in enhancing underwater imaging, which can also serve as challenging scenes for LF disparity estimation. Based on our findings from experiments, designing suitable geometric modules for underwater scenarios can enhance the performance of our paradigm, although the current depth estimation performs well hierarchically.

\bibliographystyle{IEEEtran}
\bibliography{ref}

\begin{IEEEbiography}[{\includegraphics[width=1in, height=1.25in, clip, keepaspectratio]{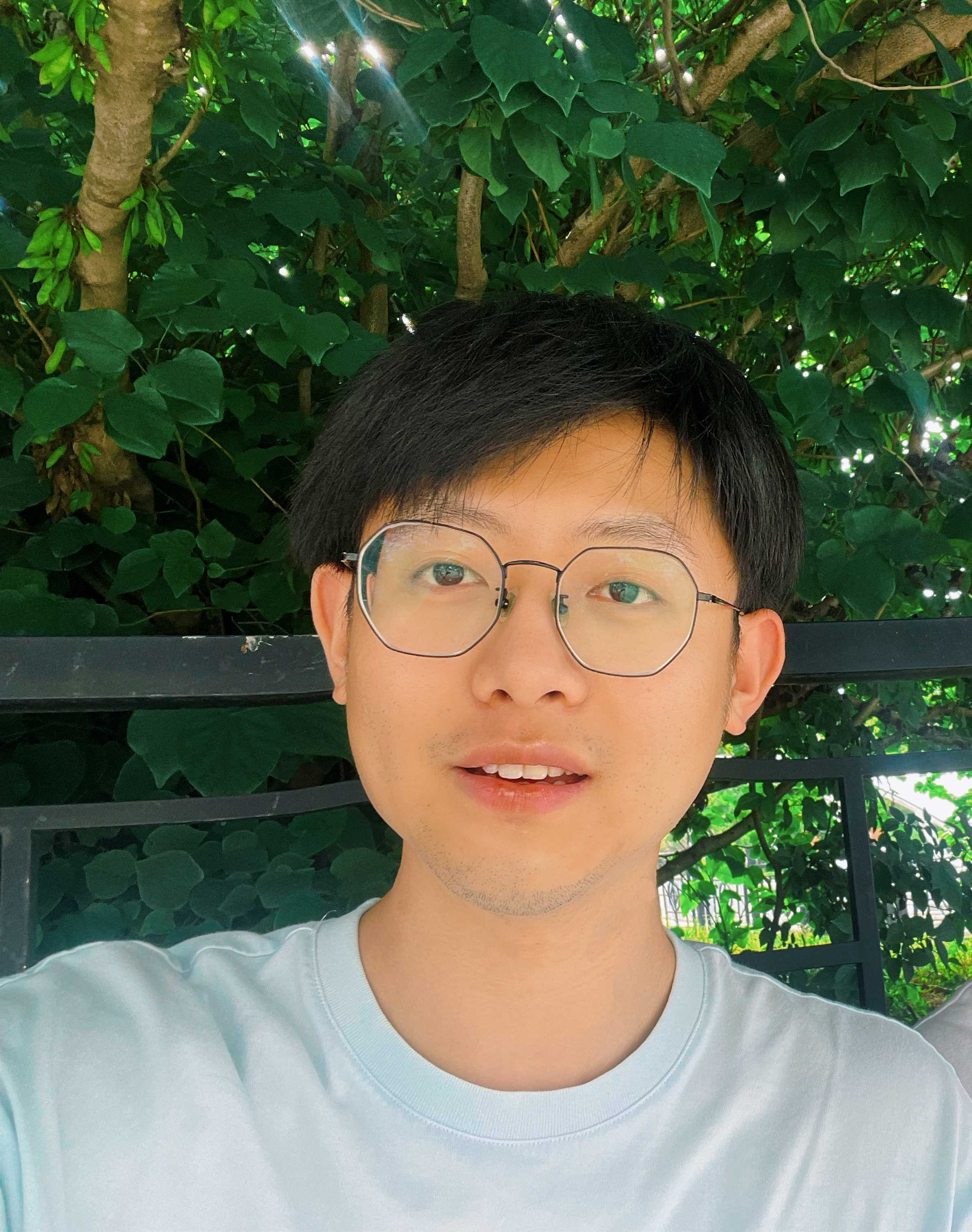}}]{Yuji Lin} received B.Eng. degrees from the School of Mathematics and Statistics, Xi'an Jiaotong University, Xi'an, China, in 2016. He is currently pursuing a Ph.D. degree with the School of Mathematics and Statistics, Xi'an Jiaotong University. His research interests include light field image processing and model-based deep learning.
\end{IEEEbiography}

\begin{IEEEbiography}[{\includegraphics[width=1in, height=1.25in, clip, keepaspectratio]{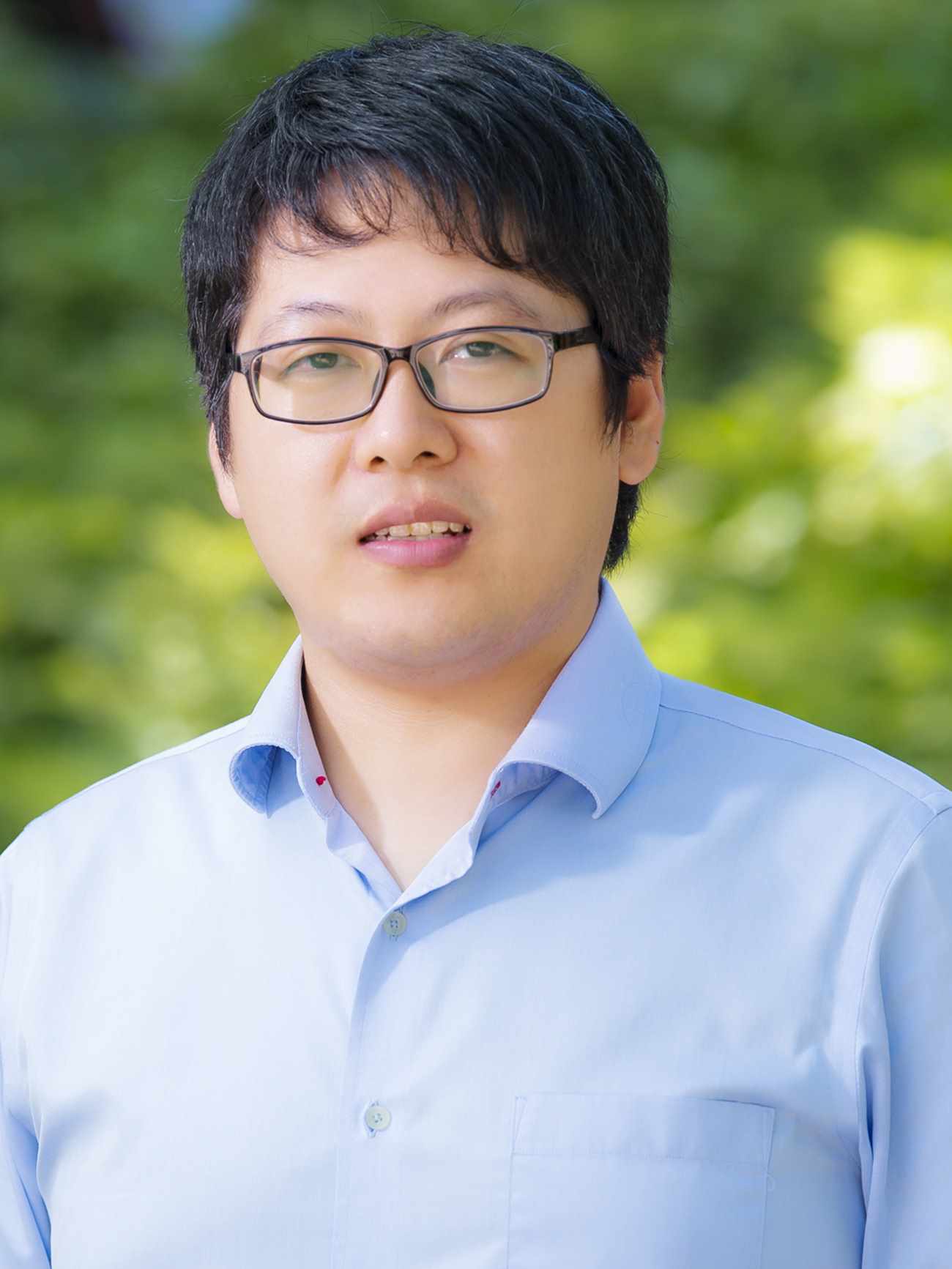}}]{Junhui Hou} (Senior Member, IEEE)  is an Associate Professor with the Department of Computer Science, City University of Hong Kong. He holds a B.Eng. degree in information engineering (Talented Students Program) from the South China University of Technology, Guangzhou, China (2009), an M.Eng. degree in signal and information processing from Northwestern Polytechnical University, Xi’an, China (2012), and a Ph.D. degree from the School of Electrical and Electronic Engineering, Nanyang Technological University, Singapore (2016). His research interests are multi-dimensional visual computing.

Dr. Hou received the Early Career Award from the Hong Kong Research Grants Council and the NSFC Excellent Young Scientists Fund. He has served or is serving as an Associate Editor for \textit{IEEE Transactions on Visualization and Computer Graphics}, \textit{IEEE Transactions on Image Processing}, \textit{IEEE Transactions on Multimedia}, and \textit{IEEE Transactions on Circuits and Systems for Video Technology}.  
\end{IEEEbiography}

\begin{IEEEbiography}[{\includegraphics[width=1in, height=1.25in, clip, keepaspectratio]{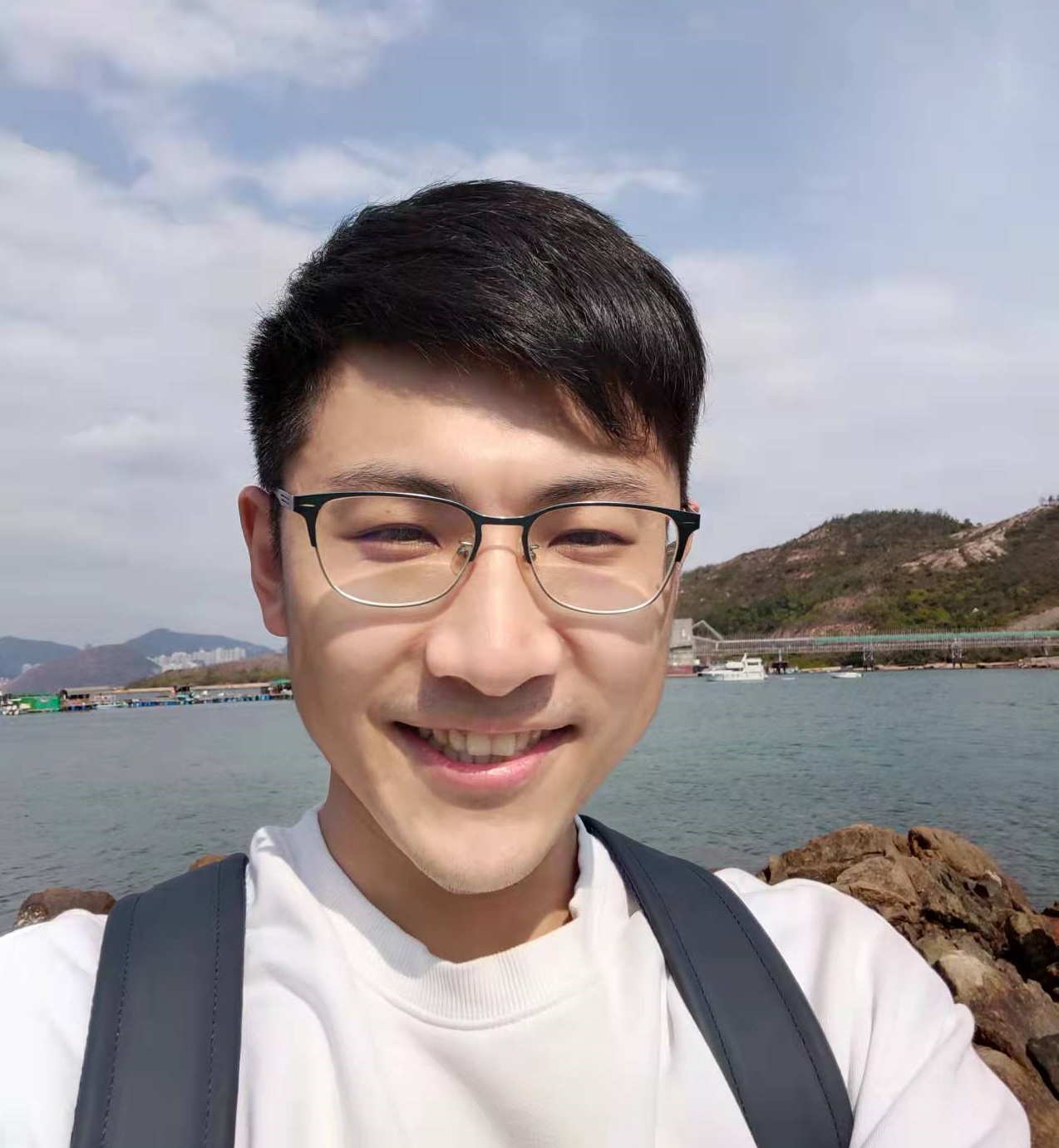}}]{Xianqiang Lyu} received B.Eng. and M.Eng. degrees from the School of Computer Science, Northwestern Polytechnical University, Xi’an, China, in 2017 and 2020, respectively, and the Ph.D. degree from the Department of Computer Science, City University of Hong Kong, in 2024. His research interests include computational photography, light field image processing, neural rendering, and deep learning.
\end{IEEEbiography}

\begin{IEEEbiography}[{\includegraphics[width=1in,height=1.25in, clip,keepaspectratio]{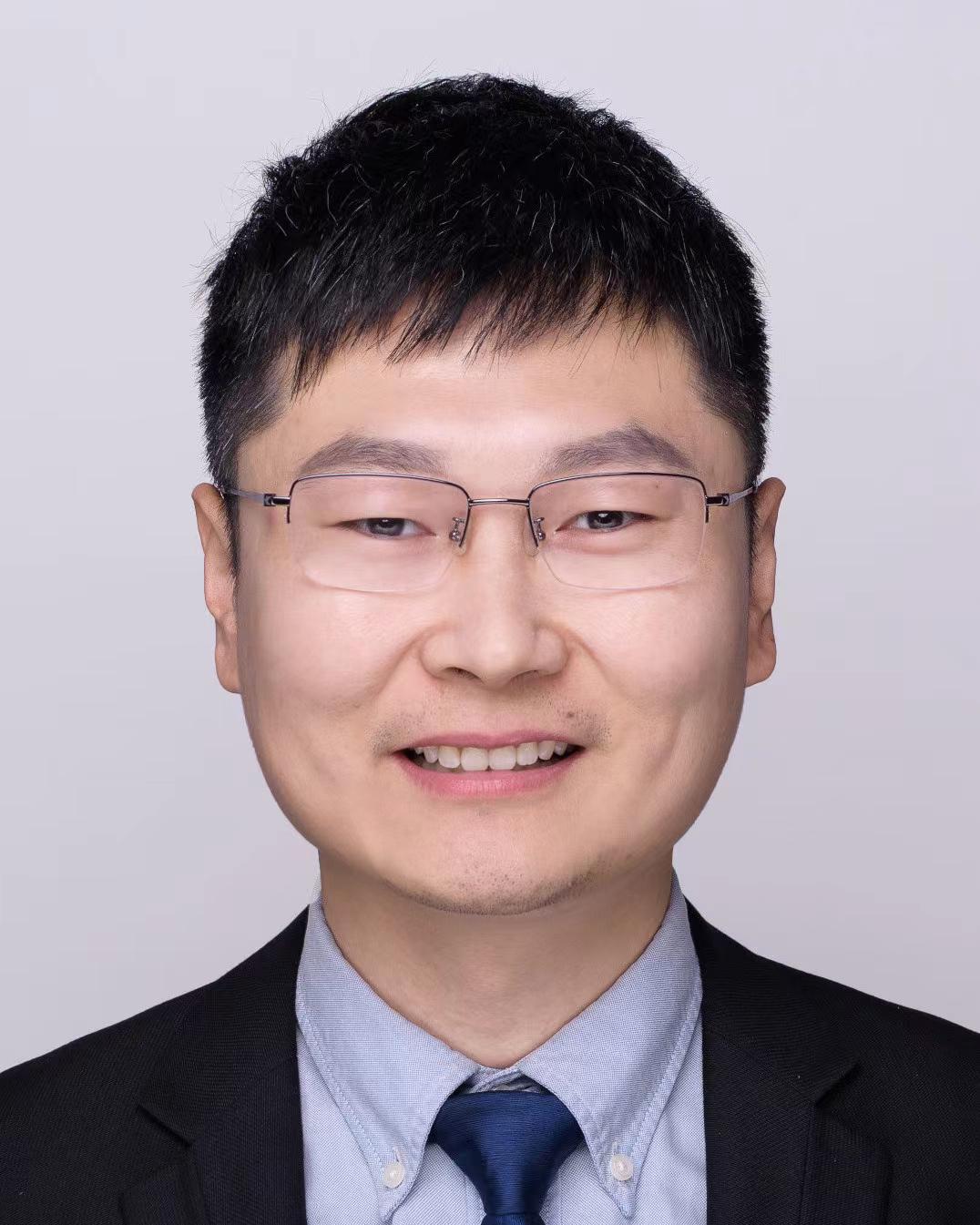}}]{Qian Zhao} (M'23) received the BSc and PhD degrees from Xi'an Jiaotong University, Xi'an, China, in 2009 and 2015, respectively. He was a visiting scholar with Carnegie Mellon University, Pittsburgh, PA, USA, from 2013 to 2014. He is currently an Associate Professor with the School of Mathematics and Statistics, Xi'an Jiaotong University. His current research interests include low-rank matrix/tensor analysis, Bayesian modeling and meta learning.
\end{IEEEbiography}

\begin{IEEEbiography}[{\includegraphics[width=1in,height=1.25in, clip,keepaspectratio]{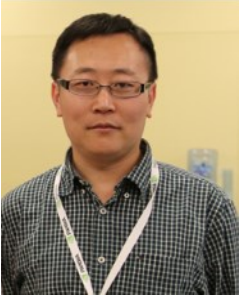}}]{Deyu Meng} received the B.Sc., M.Sc., and Ph.D. degrees from Xi'an Jiaotong University, Xi'an, China, in 2001, 2004, and 2008, respectively. He was a Visiting Scholar with Carnegie Mellon University, Pittsburgh, PA, USA, from 2012 to 2014. He is currently a professor with School of Mathematics and Statistics, Xi’an Jiaotong University, and adjunct professor with Faculty of Information Technology, The Macau University of Science and Technology. His current research interests include model-based deep learning, variational networks, and meta-learning.
\end{IEEEbiography}

\end{document}

\title{Enhancing Underwater Imaging with 4-D Light Fields: Dataset and Method \\ 
(Supplementary Materials)}

\author{Yuji Lin, Junhui Hou, \textit{Senior Member}, \textit{IEEE}, Xianqiang Lyu,  Qian Zhao, and Deyu Meng, \textit{Member}, \textit{IEEE}
}

\markboth{IEEE Journal of Selected Topics in Signal Processing}%
{Shell \MakeLowercase{\textit{et al.}}: A Sample Article Using IEEEtran.cls for IEEE Journals}


\maketitle

\section{Dataset}
\begin{figure*}[ht]
\centering
\includegraphics[width=0.9\linewidth]{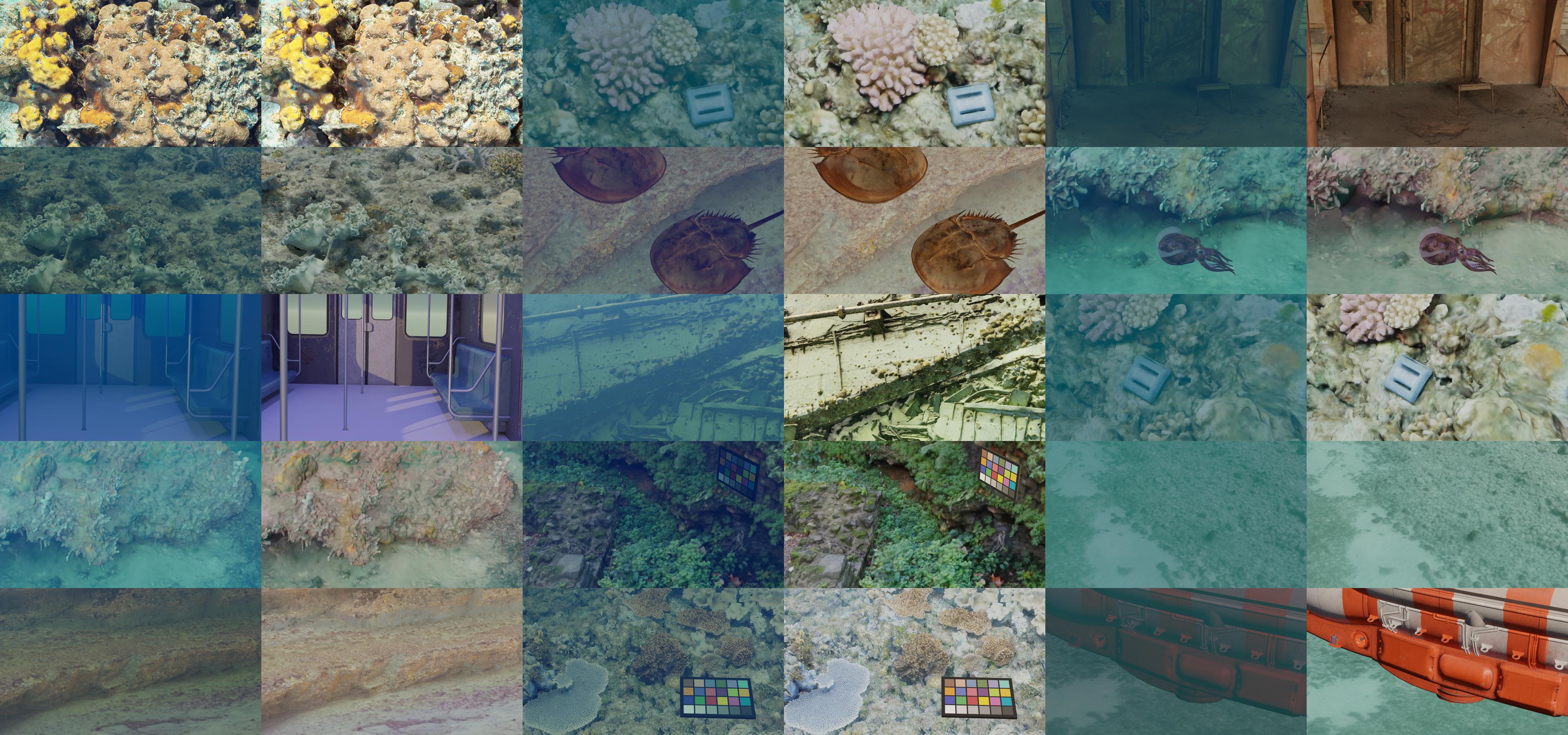}
\caption{Part one of all images.}
\label{fig:all-img-1}
\end{figure*}

\begin{figure*}[ht]
\centering
\includegraphics[width=0.9\linewidth]{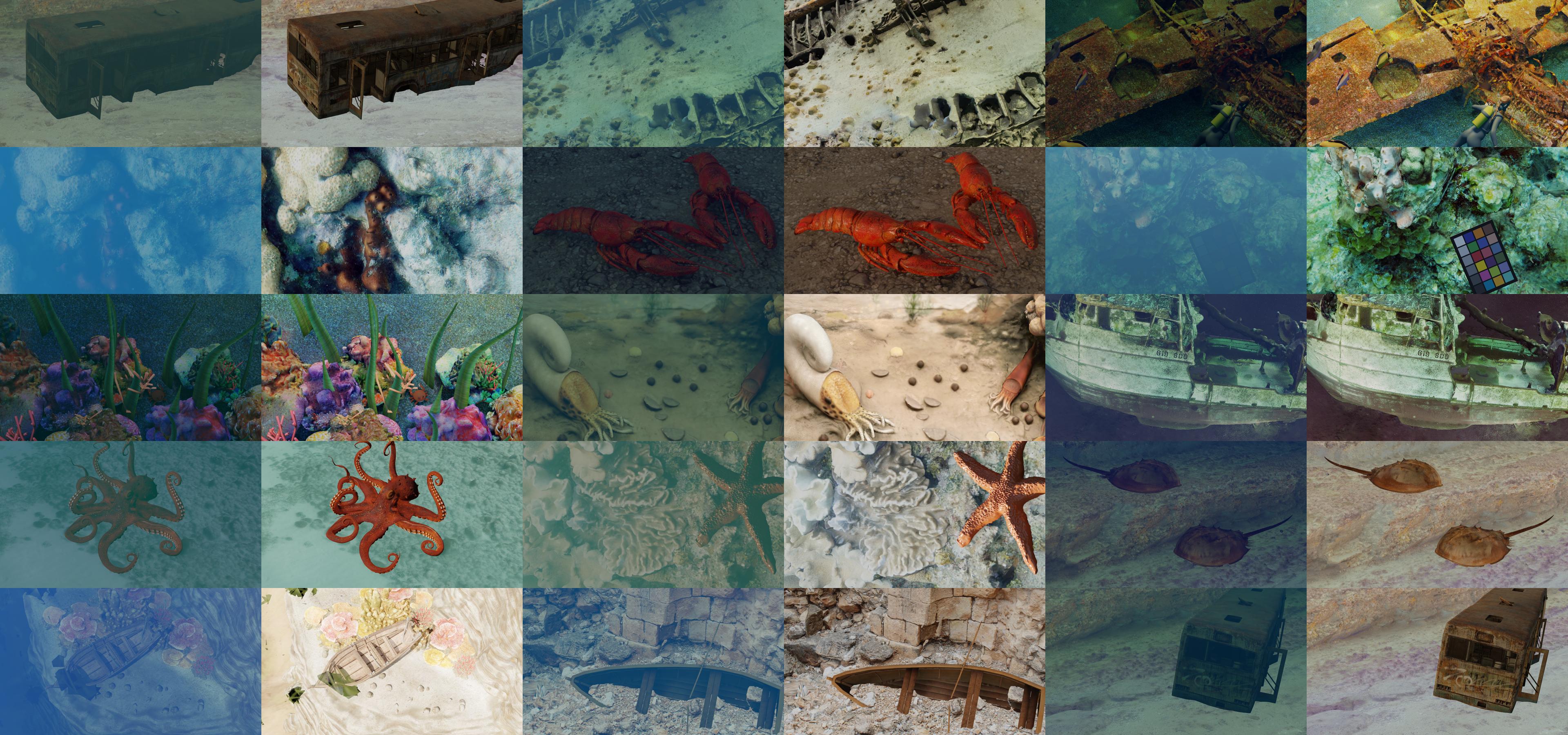}
\caption{Part two of all images.}
\label{fig:all-img-2}
\end{figure*}

\begin{figure*}[ht!]
\centering
\includegraphics[width=0.9\linewidth]{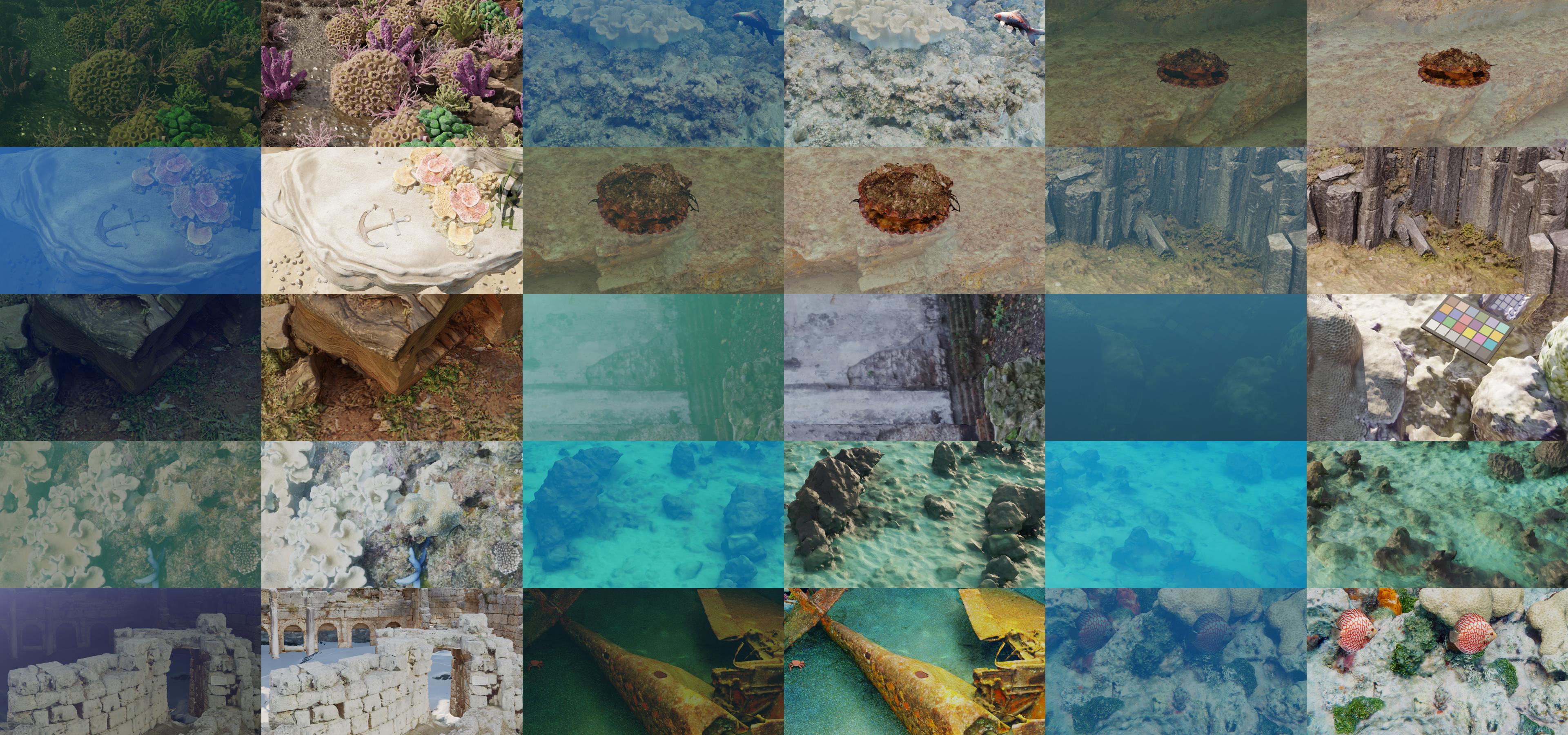}
\caption{Part three of all images.}
\label{fig:all-img-3}
\end{figure*}

\begin{figure*}[ht!]
\centering
\includegraphics[width=0.9\linewidth]{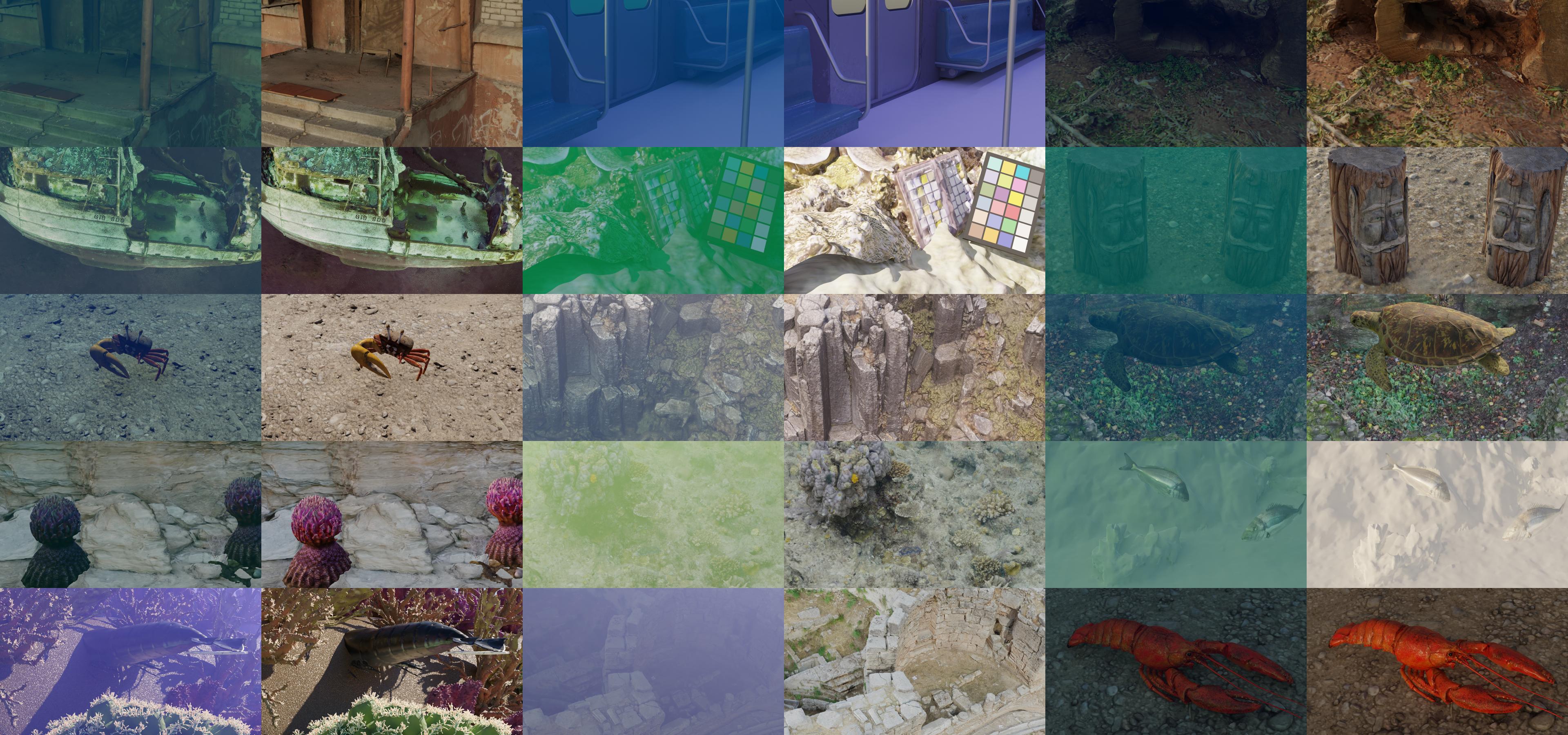}
\caption{Part four of all images.}
\label{fig:all-img-4}
\end{figure*}

We provide 70 scenes in our 4-D LF-based dataset LFUB, as shown in Fig. \ref{fig:all-img-1}, Fig. \ref{fig:all-img-2}, Fig. \ref{fig:all-img-3} and Fig. \ref{fig:all-img-4}. Each figure contains 15 scenes, including pairwise underwater and no-water center view images. 

% \vspace{-10cm}

\clearpage
\section{Experiments}
\begin{figure*}[h]
\centering
\subfloat[Input]{%
  \includegraphics[width=0.192\linewidth]{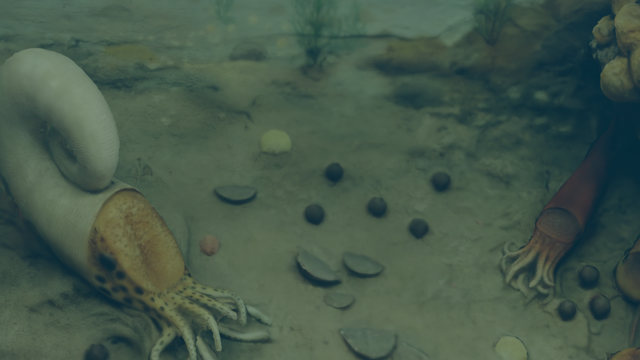}
}
\hfil
\subfloat[Fusion \cite{ancuti2012enhancing}]{%
  \includegraphics[width=0.192\linewidth]{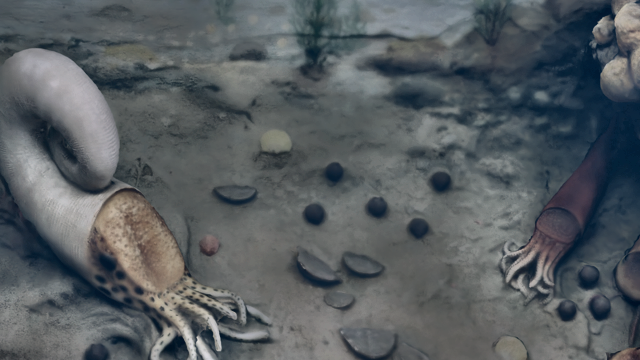}
}
\hfil
\subfloat[GDCP \cite{peng2018generalization}]{%
  \includegraphics[width=0.192\linewidth]{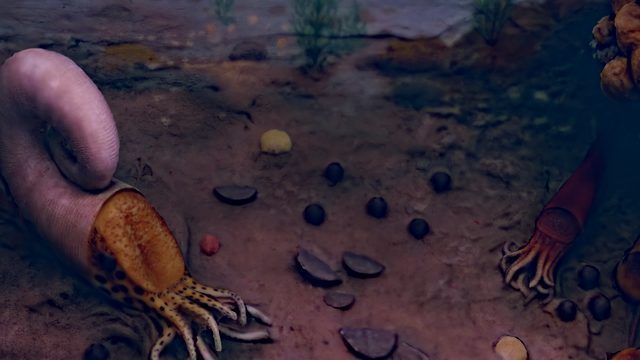}
}
\hfil
\subfloat[MMLE \cite{zhang2022underwater}]{%
  \includegraphics[width=0.192\linewidth]{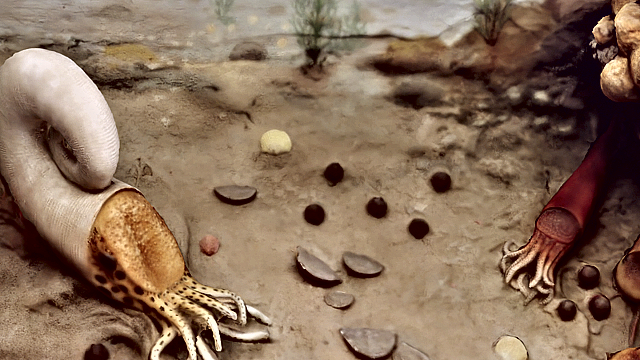}
}
\hfil
\subfloat[WWPF \cite{zhang2023underwater}]{%
  \includegraphics[width=0.192\linewidth]{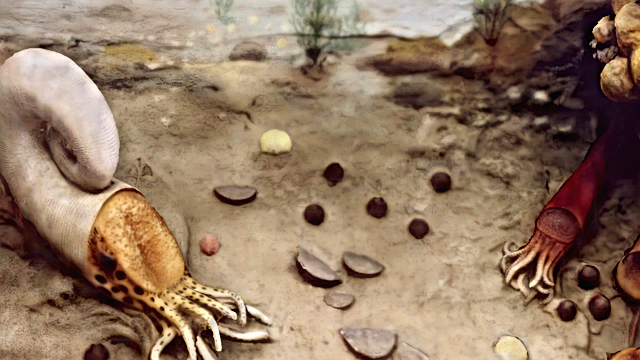}
}
\vspace{-0.3cm}

\subfloat[GT]{%
  \includegraphics[width=0.192\linewidth]{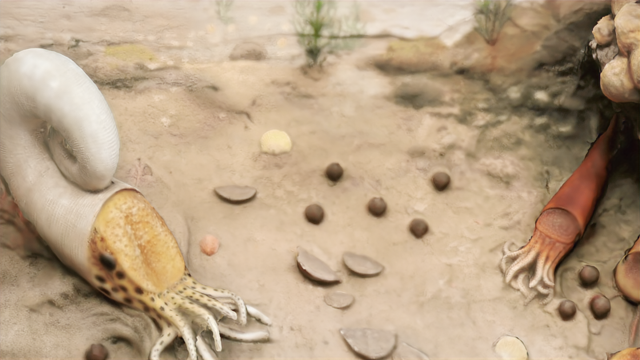}
}
\hfil
\subfloat[LANet \cite{liu2022adaptive}]{%
  \includegraphics[width=0.192\linewidth]{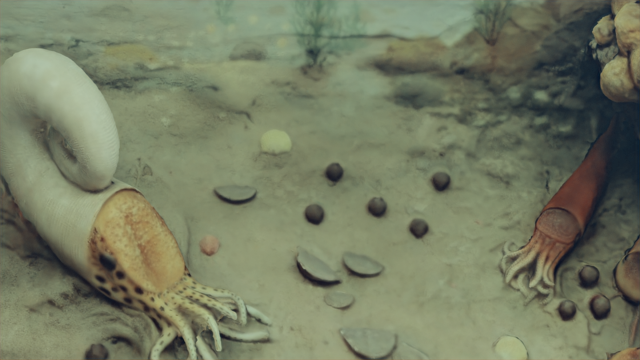}
}
\hfil
\subfloat[PUIE \cite{fu2022uncertainty}]{%
  \includegraphics[width=0.192\linewidth]{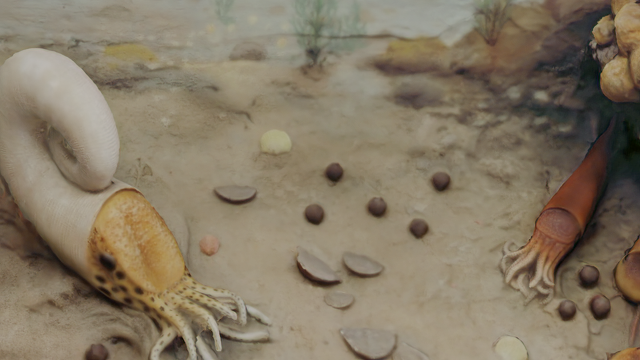}
}
\hfil
\subfloat[Ushape \cite{peng2023u}]{%
  \includegraphics[width=0.192\linewidth]{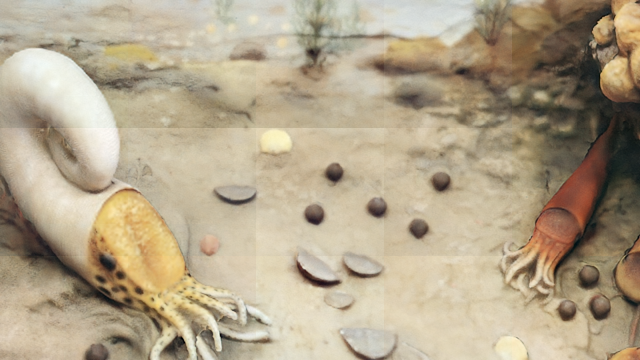}
}
\hfil
\subfloat[Uranker \cite{guo2023underwater}]{%
  \includegraphics[width=0.192\linewidth]{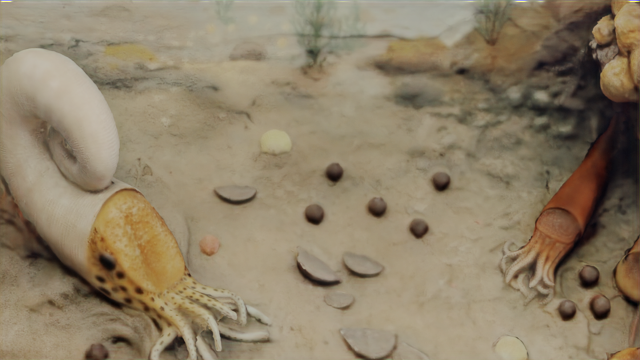}
}
\vspace{-0.3cm}

\hspace{0.1935\linewidth}
\subfloat[UVE \cite{du2024end}]{%
  \includegraphics[width=0.192\linewidth]{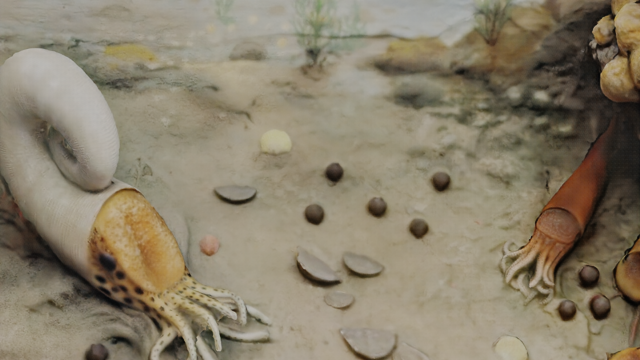}
}
\hfil
\subfloat[DistgSSR \cite{wang2022disentangling}]{%
  \includegraphics[width=0.192\linewidth]{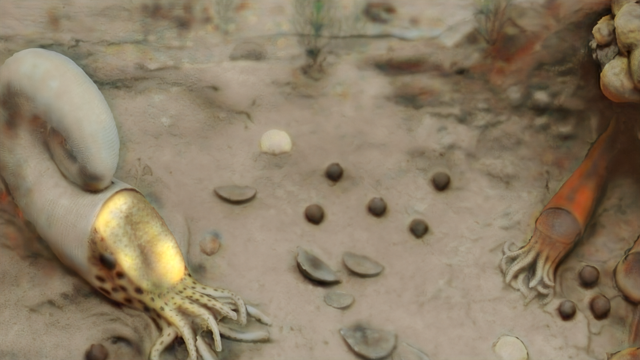}
}
\hfil
\subfloat[MSPNet \cite{wang2023multi}]{%
  \includegraphics[width=0.192\linewidth]{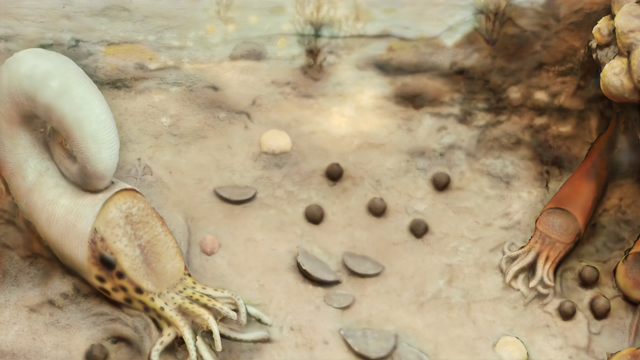}
}
\hfil
\subfloat[Ours]{%
  \includegraphics[width=0.192\linewidth]{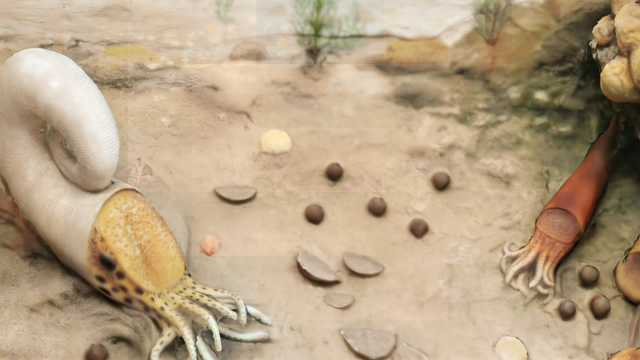}
}

\caption{Visual comparisons of different methods on \textit{Animal}.}
\label{fig:result-3}
\end{figure*}

\begin{figure*}[h]
\centering
\subfloat[Input]{%
  \includegraphics[width=0.192\linewidth]{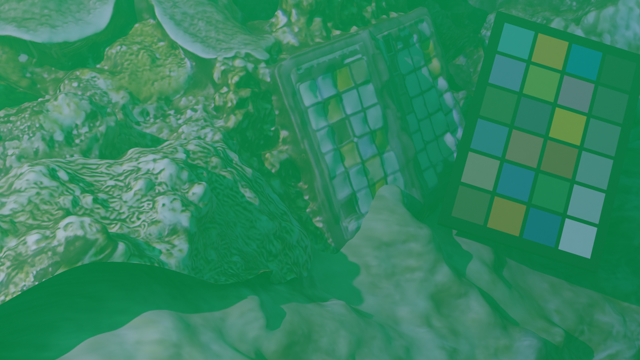}
}
\hfil
\subfloat[Fusion \cite{ancuti2012enhancing}]{%
  \includegraphics[width=0.192\linewidth]{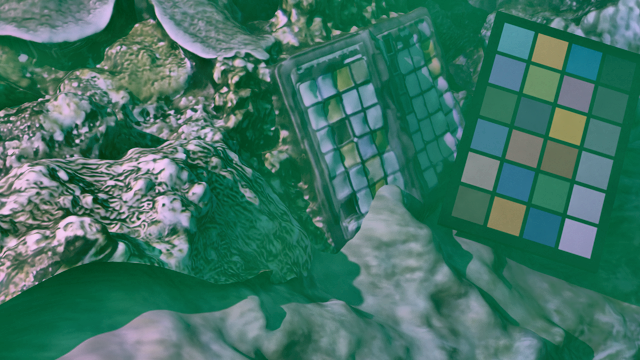}
}
\hfil
\subfloat[GDCP \cite{peng2018generalization}]{%
  \includegraphics[width=0.192\linewidth]{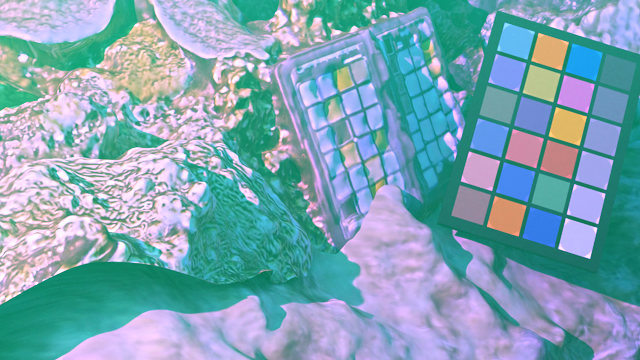}
}
\hfil
\subfloat[MMLE \cite{zhang2022underwater}]{%
  \includegraphics[width=0.192\linewidth]{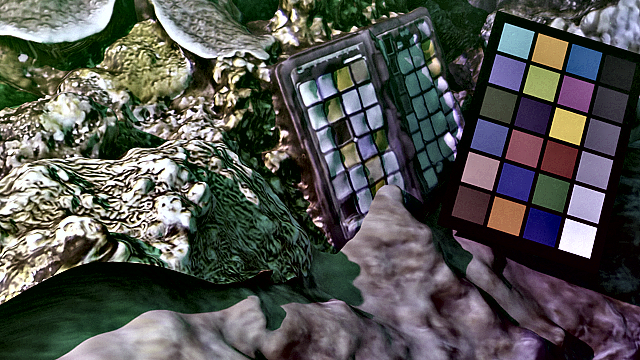}
}
\hfil
\subfloat[WWPF \cite{zhang2023underwater}]{%
  \includegraphics[width=0.192\linewidth]{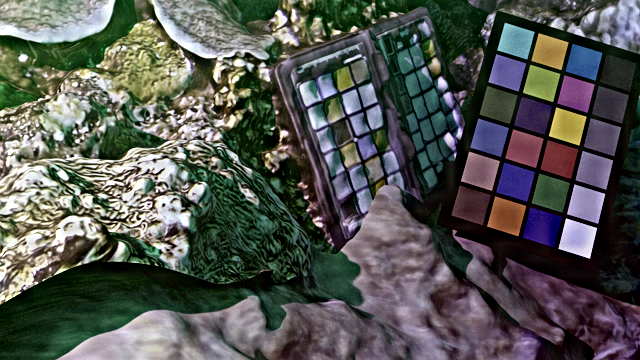}
}
\vspace{-0.3cm}

\subfloat[GT]{%
  \includegraphics[width=0.192\linewidth]{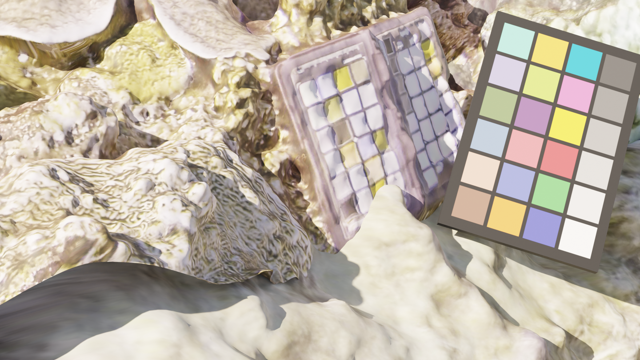}
}
\hfil
\subfloat[LANet \cite{liu2022adaptive}]{%
  \includegraphics[width=0.192\linewidth]{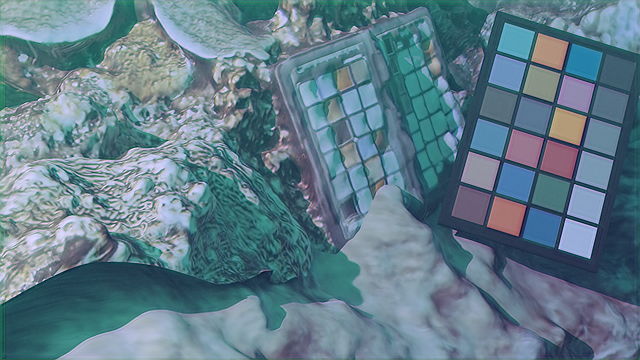}
}
\hfil
\subfloat[PUIE \cite{fu2022uncertainty}]{%
  \includegraphics[width=0.192\linewidth]{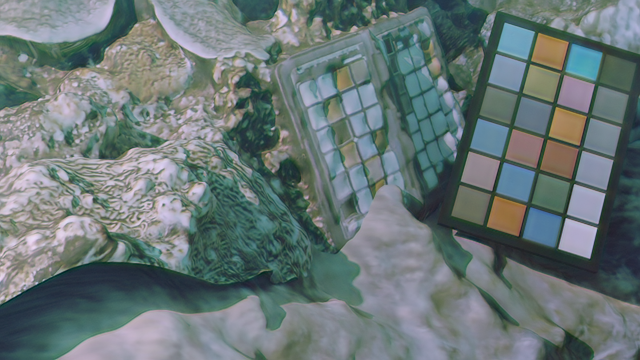}
}
\hfil
\subfloat[Ushape \cite{peng2023u}]{%
  \includegraphics[width=0.192\linewidth]{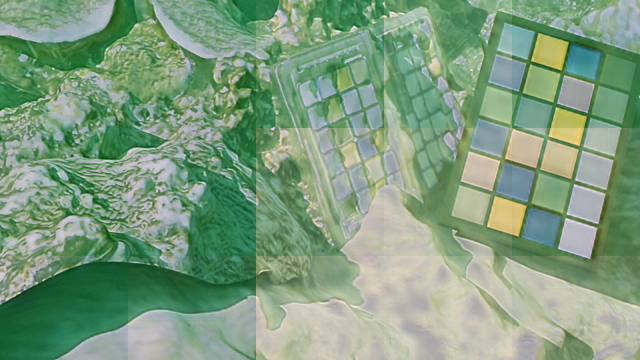}
}
\hfil
\subfloat[Uranker \cite{guo2023underwater}]{%
  \includegraphics[width=0.192\linewidth]{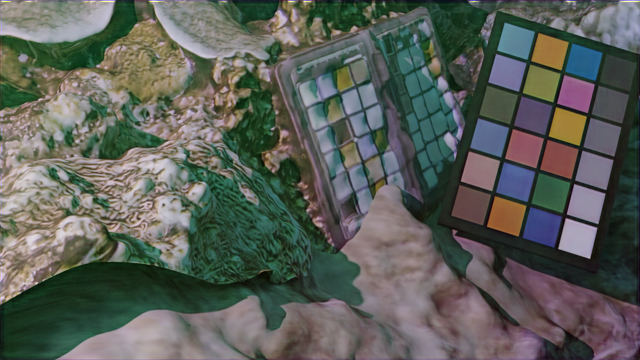}
}
\vspace{-0.3cm}

\hspace{0.1935\linewidth}
\subfloat[UVE \cite{du2024end}]{%
  \includegraphics[width=0.192\linewidth]{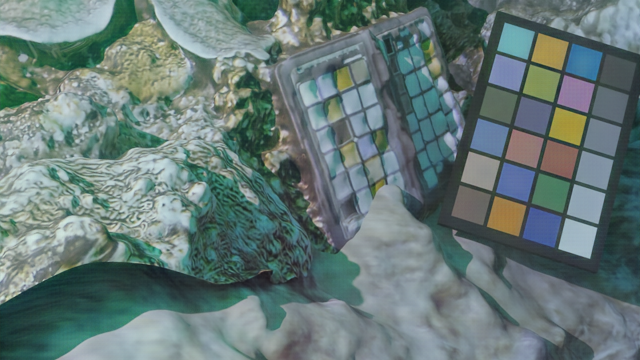}
}
\hfil
\subfloat[DistgSSR \cite{wang2022disentangling}]{%
  \includegraphics[width=0.192\linewidth]{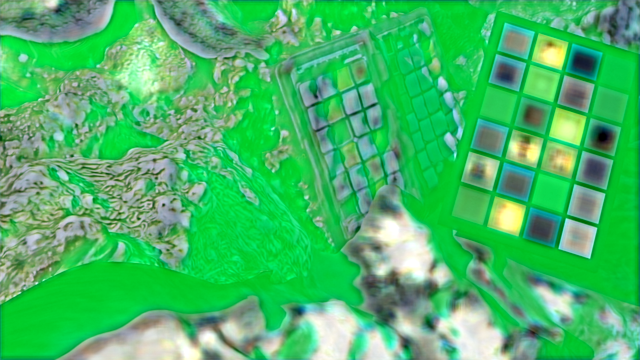}
}
\hfil
\subfloat[MSPNet \cite{wang2023multi}]{%
  \includegraphics[width=0.192\linewidth]{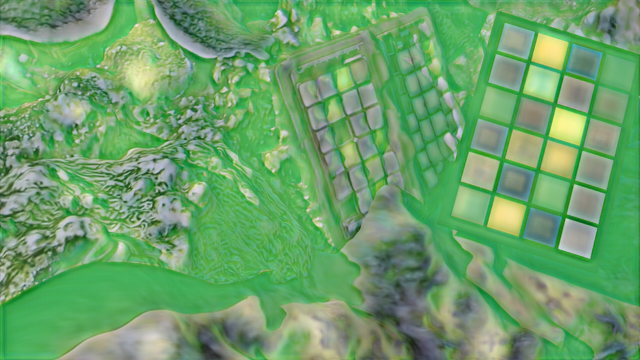}
}
\hfil
\subfloat[Ours]{%
  \includegraphics[width=0.192\linewidth]{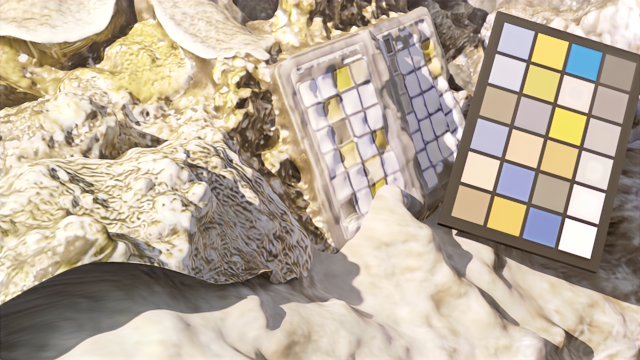}
}

\caption{Visual comparisons of different methods on \textit{Color Checker}.}
\label{fig:result-4}
\end{figure*}

We provide more enhancement results in Fig. \ref{fig:result-3} and Fig. \ref{fig:result-4}. 
We show the comparisons on the scene \textit{Animal} with greenish color deviation in Fig. \ref{fig:result-3}. Although most methods are able to restore the correct colors, only a few, including Ushape, MSPNet, and Ours, achieve normal contrast. But Ushape and MSPNet introduce extra color artifacts, while ours achieve the best enhancement results. We show a scene with a color-checker with severe greenish color deviation in Fig. \ref{fig:result-4}. Almost all methods fail to restore the correct colors, with the overall scene still exhibiting a green hue. However, our method is able to completely remove the green from the scene. Unfortunately, the colors on the color checker are overly removed, likely due to the uniform application of depth information. This is one of the directions we aim to improve in the future.

\clearpage

\begin{figure*}[h]
    \centering
    \begin{tikzpicture}
        \node[anchor=south west,inner sep=0] (image) at (0,0) {\includegraphics[width=0.9\textwidth]{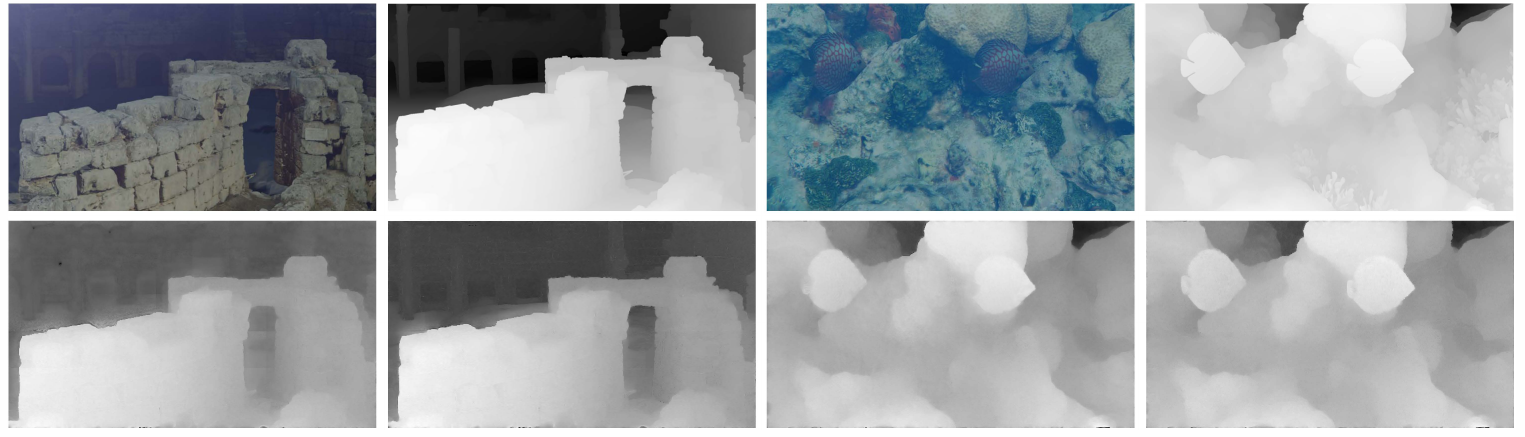}};
        \node[align=center, rotate=90, font=\small] at (-0.1,3.3) {Input};
        \node[align=center, rotate=90, font=\small] at (-0.1,1.15) {OccUnNet};
        \node[align=center, font=\small] at (2,4.9) {\textit{Statue}};
        \node[align=center, font=\small] at (6.1,4.9) {Depth GT};
        \node[align=center, font=\small] at (10.2,4.9) {\textit{Coral}};
        \node[align=center, font=\small] at (14.2,4.9) {Depth GT};
        \node[align=center, font=\small] at (2,-0.15) {Depth for Input};
        \node[align=center, font=\small] at (6.1,-0.15) {Depth for Result};
        \node[align=center, font=\small] at (10.2,-0.15) {Depth for Input};
        \node[align=center, font=\small] at (14.2,-0.15) {Depth for Result};
    \end{tikzpicture}
    \caption{Disparity maps estimated by OccUnNet on proposed LFUB dataset. Top row: (from left to right) degraded \textit{Statue} image, \textit{Statue} GT Depth map, degraded \textit{Coral} image, \textit{Coral} GT Depth map. Bottom row: disparity maps estimated by OccUnNet of the degraded and restored scenes.} 
    \label{fig:depth2}
\end{figure*}

We provide more depth estimation results in Fig. \ref{fig:depth2}.

\clearpage
\bibliographystyle{IEEEtran}
\bibliography{ref}